\documentclass{article} 
\usepackage{iclr2020_conference,times}

\usepackage{hyperref}
\usepackage{url}
\usepackage{epsfig}
\usepackage{graphicx}
\usepackage{amsmath}
\usepackage{amssymb}
\usepackage{epsfig}
\usepackage{graphicx}
\usepackage{amsmath}
\usepackage{amssymb}
\usepackage{amsfonts}    
\usepackage{nicefrac}       
\usepackage{microtype}      
\usepackage{xcolor}
\usepackage{grffile}
\usepackage{amsmath}
\usepackage{graphicx}
\usepackage{verbatim}
\usepackage{setspace}
\usepackage{subcaption}
\usepackage{verbatim}
\usepackage{multirow}
\usepackage{mwe}
\usepackage{verbatimbox}
\usepackage{array,booktabs}
\usepackage[inline]{enumitem}
\usepackage{mathtools}
\usepackage{floatrow}
\usepackage{multirow}
\usepackage{placeins}
\usepackage{array}
\usepackage{hypcap}
\usepackage{subcaption}
\usepackage{capt-of}
\usepackage{arydshln}
\usepackage{wrapfig}
\usepackage[para]{footmisc}

\newcommand{\myparagraph}[1]{\vspace{0.0em}\noindent\textbf{#1}}
\newcolumntype{C}[1]{>{\centering\let\newline\\\arraybackslash\hspace{0pt}}m{#1}}

\definecolor{christophcolor}{rgb}{0,0,1}
\newcommand{\christoph}[1]{}

\definecolor{berntcolor}{rgb}{0.6,0.4,0.8}
\newcommand{\bernt}[1]{}

\definecolor{mariocolor}{rgb}{1,0,1}
\newcommand{\mario}[1]{}
\newcommand{\marioo}[1]{}

\iclrfinalcopy 

\begin{document}
\title{Conditional Flow Variational Autoencoders for Structured Sequence Prediction}

\author{Apratim Bhattacharyya \thanks{\tiny{Max Planck Institute for Informatics, Saarbr\"{u}cken, Germany,}} \quad Michael Hanselmann \thanks{\tiny{Bosch Center for Artificial Intelligence, Renningen, Germany,}} \quad Mario Fritz \thanks{\tiny{CISPA Helmholtz Center for Information Security, Saarbr\"{u}cken, Germany}} \quad Bernt Schiele \footnotemark[1] \\ Christoph-Nikolas Straehle \footnotemark[2]}

\newcommand{\fix}{\marginpar{FIX}}
\newcommand{\new}{\marginpar{NEW}}
\maketitle

\vspace{-1cm}
\begin{abstract}
Prediction of future states of the environment and interacting agents is a key competence required for autonomous agents to operate successfully in the real world.
Prior work for structured sequence prediction based on latent variable models imposes a uni-modal standard Gaussian prior on the latent variables. This induces a strong model bias which makes it challenging to fully capture the multi-modality of the distribution of the future states.
In this work, we introduce \emph{Conditional Flow Variational Autoencoders (CF-VAE)} using our novel conditional normalizing flow based prior to capture complex multi-modal conditional distributions for effective structured sequence prediction. Moreover, we propose two novel regularization schemes which stabilizes training and deals with posterior collapse for stable training and better fit to the target data distribution.
Our experiments on three multi-modal structured sequence prediction datasets -- MNIST Sequences, Stanford Drone and HighD -- show that the proposed method obtains state of art results across different evaluation metrics.

\end{abstract}

\mario{the text uses ``over-regularization''. not sure, if I like the term. maybe rather ``bias'' or ``induce a strong model bias''?}



\section{Introduction}
\mario{general comment: In my opinion "xyz based" should be changed to "xyz-based". E.g. "flow based" to "flow-based" }
Anticipating future states of the environment is a key competence necessary for the success of autonomous agents.
In complex real world environments, the future is highly uncertain.
Therefore, structured predictions, one to many mappings \citep{sohn2015learning,bhattacharyya2018accurate} of the likely future states of the world, are important.
In many scenarios, these tasks can be cast as sequence prediction problems. Particularly, Conditional Variational Autoencoders (CVAE) \citep{sohn2015learning} have been successful for such problems -- from prediction of future pedestrians trajectories \citep{lee2017desire,bhattacharyya2018accurate,pajouheshgar2018back} to outcomes of robotic actions \citep{babaeizadeh2017stochastic}. The distribution of future sequences is diverse and highly multi-modal.
CVAEs model diverse futures by factorizing the distribution of future states using a set of latent variables which are mapped to likely future states. 
However, CVAEs assume a standard Gaussian prior on the latent variables which induces a strong model bias \citep{hoffman2016elbo,tomczak2017vae} which makes it challenging to capture multi-modal distributions. This also leads to missing modes due to posterior collapse \citep{bowman2015generating,razavi2019preventing}.

Recent work \citep{tomczak2017vae,wang2017diverse,gu2018dialogwae} has therefore focused on more expressive Gaussian mixture based priors. However, Gaussian mixtures still have limited expressiveness and optimization suffers from complications e.g.\ determining the number of mixture components. In contrast, normalizing flows are more expressive and enable the modelling of complex multi-modal priors. Recent work on flow based priors \citep{chen2016variational,ziegler2019latent}, have focused only on the unconditional (plain VAE) case. However, this not sufficient for CVAEs because in the conditional case the complexity of the distributions are highly dependent on the condition. 

\begin{figure}[!t]
\centering
	\begin{tabular}{cccccc}
	\toprule
	{Latent Prior} & \multicolumn{2}{c}{Clustered Predictions} & {Latent Prior} & \multicolumn{2}{c}{{Clustered Predictions}} \\
	\midrule
	\includegraphics[height=1.4cm,width=1.4cm,trim={1.5cm 1cm 1cm 0.5cm},clip]{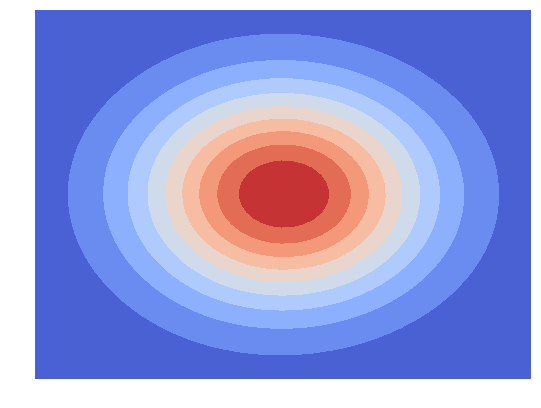} &
	\includegraphics[height=1.4cm,width=1.4cm,trim={2cm 0.8cm 0.8cm 1.5cm},clip]{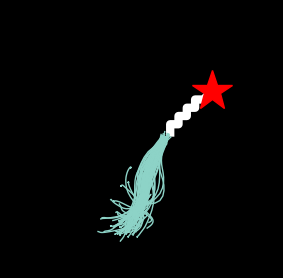} &
	\includegraphics[height=1.4cm,width=1.4cm,trim={2cm 0.8cm 0.8cm 1.5cm},clip]{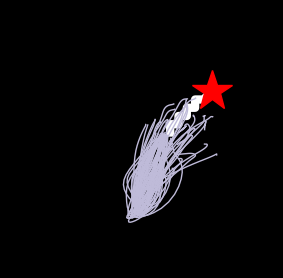} &
	\includegraphics[height=1.4cm,width=1.4cm,trim={1.5cm 1cm 1cm 0.5cm},clip]{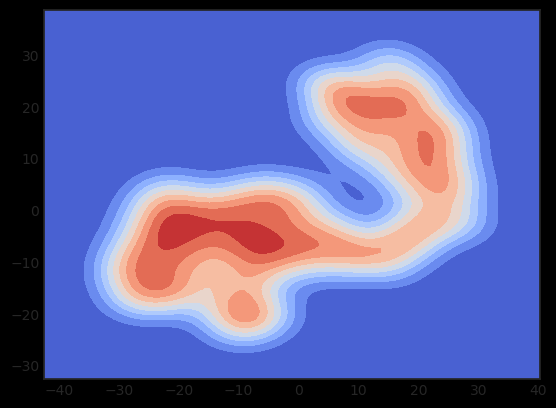} &
	\includegraphics[height=1.4cm,width=1.4cm,trim={2cm 0.8cm 0.8cm 1.5cm},clip]{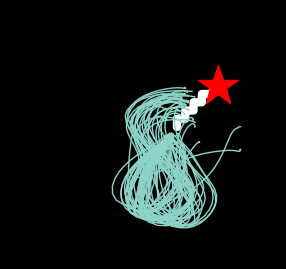} &
	\includegraphics[height=1.4cm,width=1.4cm,trim={2cm 0.8cm 0.8cm 1.5cm},clip]{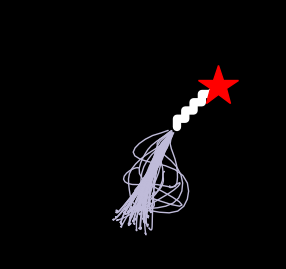} \\
	\multicolumn{3}{c}{\upbracefill}& \multicolumn{3}{c}{\upbracefill}\\
	\multicolumn{3}{c}{Standard Gaussian Prior -- CVAE.}  & \multicolumn{3}{c}{Conditional Flow Prior -- Our CF-VAE.} \\
	\bottomrule
	\end{tabular}
\caption{Clustered stroke predictions on MNIST sequences. Our multi-modal Conditional Normalizing Flow based prior (right) enables our regularized CF-VAE to capture the two modes of the conditional distribution, while predictions with uni-modal Gaussian prior (left) have limited diversity. Note, our 64D CF-VAE latent distribution is (approximately) projected to 2D using tSNE and KDE.}
\label{fig:teaser}
\end{figure}
In this work, \begin{enumerate*} \item We propose \emph{Conditional Flow Variational Autoencoders (CF-VAE)} based on novel conditional normalizing flow based priors In order to model complex multi-modal conditional distributions over sequences. In \autoref{fig:teaser}, we show example predictions of MNIST handwriting stroke of our CF-VAE. We observe that, given a starting stroke, our CF-VAE model with data dependent normalizing flow based latent prior captures the two main modes of the conditional distribution -- i.e. 1 and 8 -- while CVAEs with fixed uni-modal Gaussian prior predictions have limited diversity. \item  We propose a novel regularization scheme that stabilizes the optimization of the evidence lower bound and leads to better fit to the target data distribution. \item We leverage our conditional flow prior to deal with posterior collapse which causes standard CVAEs to ignore modes in sequence prediction tasks. \item Finally, our method outperforms the state of the art on three structured sequence prediction tasks -- handwriting stroke prediction on MNIST, trajectory prediction on Stanford Drone and HighD.  \end{enumerate*}


\section{Related Work}
\label{sec:relatedwork}
\myparagraph{Normalizing Flows.} Normalizing flows are a powerful class of density estimation methods with exact inference. \citep{dinh2014nice} introduced affine normalizing flows with triangular Jacobians. \citep{dinh2016density} extend flows with masked convolutions which allow for complex (non-autoregessive) dependence between the dimensions. In \citep{kingma2018glow}, $1 \times 1$ convolutions were proposed for improved image generation compared to \citep{dinh2016density}.  
In \citep{huang2018neural} normalizing flows are auto-regressive and \citep{behrmann2018invertible} extend it to ResNet.
\citep{lu2019structured} extended normalizing flows to model conditional distributions.
Here, we propose conditional normalizing flows to learn conditional priors for variational latent models.

\myparagraph{Variational Autoencoders.}  The original variational autoencoder \citep{kingma2013auto} used uni-modal Gaussian prior and posterior distributions. Thereafter, two lines of work have focused on developing either more expressive prior or posterior distributions.
\cite{rezende2015variational} propose normalizing flows to model complex posterior distributions. 
\cite{kingma2016improved,tomczak2016improving,berg2018sylvester} present more complex inverse autoregessive flows, householder and Sylvester normalizing flow based posteriors. Here, we focus on the orthogonal direction of more expressive priors and the above approaches are compatible with our approach.

Recent work which focus more expressive priors include \citep{nalisnick2016stick} which proposes a Dirichlet process prior and \citep{goyal2017nonparametric} which proposes a nested Chinese restaurant process prior. However, these methods require sophisticated learning methods.
In contrast, \citep{tomczak2017vae} proposes a mixture of Gaussians based prior (with fixed number of components) which is easier to train and shows promising results on some image generation tasks. \citep{chen2016variational}, proposes a inverse autoregressive flow based prior which leads to improvements in complex image generation tasks like CIFAR-10. 
\citep{ziegler2019latent} proposes a prior for VAE based text generation using complex non-linear flows which allows for complex multi-modal priors. 
While these works focus on unconditional priors, we aim to develop more expressive conditional priors. 
 
\myparagraph{Posterior Collapse.} Posterior collapse arises when the latent posterior does not encode useful information. Most prior work \citep{yang2017improved,dieng2018avoiding,higgins2017beta} concentrate on unconditional VAEs and modify the training objective -- the KL divergence term is annealed to prevent collapse to the prior. \cite{liu2019cyclical} extends KL annealing to CVAEs. However, KL annealing does not optimize a true lower bound of the ELBO for most of training. \cite{zhao2017infovae} also modifies the objective to choose the model with the maximal rate.  \cite{razavi2019preventing} propose anti-causal sequential priors for text modelling tasks. \cite{bowman2015generating,gulrajani2016pixelvae} proposes to weaken the decoder so that the latent variables cannot be ignored, however only unconditional VAEs are considered. \cite{wang2019riemannian} shows the advantage of normalizing flow based posteriors for preventing posterior collapse. In contrast, we study for the first time posterior collapse in conditional models on datasets with minor modes.

\myparagraph{Structured Sequence Prediction.} \cite{helbing1995social,robicquet2016learning,alahi2016social,gupta2018social,zhao2019multi,sadeghian2018sophie} consider the problem of traffic participant trajectory prediction in a social context. Notably, \citep{gupta2018social,zhao2019multi,sadeghian2018sophie} use generative adversarial networks to generate socially compliant trajectories. However, the predictions are uni-modal. \cite{lee2017desire,bhattacharyya2018accurate,rhinehart2018r2p2,deo2019scene,pajouheshgar2018back} considers structured (one to many) predictions using  -- a CVAE, improved CVAE training, pushforward policies for vehicle ego-motion prediction, motion planning, spatio-temporal convolutional network respectively. \cite{kumar2019videoflow} proposes a normalizing flow based model for video sequence prediction, however the sequences considered have very limited diversity compared to the trajectory prediction tasks considered here.  Here, we focus on improving structured predictions using conditional normalizing flows based priors. 


\section{Conditional Flow Variational Autoencoder (CF-VAE)}
\label{sec:cvae}
Our Conditional Flow Variational Autoencoder is based on the conditional variational autoencoder \citep{sohn2015learning} which is a deep directed graphical model for modeling conditional data distributions $p_{\theta}(\text{y}|\text{x})$. Here, $\text{x}$ is the sequence up to time $t$, $x = \left[ x^{1}, \cdots,  x^{t} \right]$ and $\text{y}$ is the sequence to be predicted up to time $T$, $y = \left[ y^{t+1}, \cdots,  y^{T} \right]$. CVAEs factorize the conditional distribution using latent variables $\text{z}$ -- $p_{\theta}(\text{y}|\text{x})$ is factorized as $p_{\theta}(\text{y}|\text{z}, \text{x}) p(\text{z}|\text{x})$, where $p(\text{z}|\text{x})$ is the prior on the latent variables. During training, amortized variational inference is used and the posterior distribution $q_{\phi}(\text{z}|\text{x},\text{y})$ is learnt using a recognition network. The ELBO is maximized, given by,
\begin{align}\label{eq:cvae}
\log(p_{\theta}(\text{y}|\text{x})) \geq \mathbb{E}_{q_{\phi}(\text{z}|\text{x},\text{y})} \log(p_{\theta}(\text{y}|\text{z},\text{x})) - D_{\text{KL}}( q_{\phi}(\text{z}|\text{x},\text{y}) || p(\text{z}|\text{x})).  
\end{align}
In practice, to simplify learning, simple unconditional standard Gaussian priors are used \citep{sohn2015learning}. However, the complexity e.g.\ the number of modes of the target distributions $p_{\theta}(\text{y}|\text{x})$, is highly dependent upon the condition $x$. An unconditional prior demands identical latent distributions irrespective complexity of the target conditional distribution -- a very strong constraint on the recognition network. Moreover, the latent variables cannot encode any conditioning information and this leaves the burden of learning the dependence on the condition completely on the decoder. 

Furthermore, on complex conditional multi-modal data, Gaussian priors have been shown to induce a strong model bias \citep{tomczak2016improving,ziegler2019latent}. It becomes increasingly difficult to map complex multi-modal distributions to uni-modal Gaussian distributions, further complicated by the sensitivity of the RNNs encoder/decoders to subtle variations in the hidden states \citep{bowman2015generating}. Moreover, the standard closed form estimate of the KL-divergence pushes the encoded latent distributions to the mean of the Gaussian leading to latent variable collapse \citep{wang2017diverse,gu2018dialogwae} while discriminator based approaches \citep{tolstikhin2017wasserstein} lead to underestimates of the KL-divergence \citep{rosca2017variational}.  \marioo{too strong? i made the statment a bit stronger - and connected the two sentences - as the last one was dangling a bit.   }

Therefore, we propose conditional priors based on conditional normalizing flows to enable the latent variables to encode conditional information and allow for complex multi-modal latent representations. Next, we introduce our novel conditional non-linear normalizing flows followed by our novel regularized Conditional Flow Variational Autoencoder (CF-VAE) formulation.


\subsection{Conditional Normalizing Flows}
Recently, normalizing flow \citep{tabak2010density,dinh2014nice} based priors for VAEs have been proposed \citep{chen2016variational,ziegler2019latent}.
Normalizing flows allows for complex priors by transforming a simple base density e.g.\ standard Gaussian to a complex multi-modal density through a series of $n$ layers of invertible transformations $f_{i}$,
\begin{align}\label{eq:unflowtrans}
\epsilon \overset{f_{1}}{\longleftrightarrow} \text{h}_{1} \overset{f_{2}}{\longleftrightarrow} \text{h}_{2} \cdots \overset{f_{n}}{\longleftrightarrow} \text{z}.
\end{align}

However, such flows cannot model conditional priors. In contrast to prior work, we utilize conditional normalizing flows to model complex conditional priors. Conditional normalizing flows also consists of a series of $n$ layers of invertible transformations $f_{i}$ (with parameters $\psi$), however we modify the transformations $f_{i}$ such that they are dependent on the condition $\text{x}$,
\begin{align}\label{eq:flowtrans}
\epsilon | \text{x} \overset{f_{1} | \text{x} }{\longleftrightarrow} \text{h}_{1} | \text{x} \overset{f_{2} | \text{x} }{\longleftrightarrow} \text{h}_{2} | \text{x} \cdots \overset{f_{n} | \text{x} }{\longleftrightarrow} \text{z} | \text{x}.
\end{align}

Further, in contrast to prior work \citep{lu2019structured,atanov2019semi,ardizzone2018analyzing} which use affine flows ($f_{i}$), we build upon \citep{ziegler2019latent} and introduce conditional non-linear normalizing flows with split coupling. Split couplings ensure invertibility by applying a flow layer $f_{i}$ on only half of the dimensions at a time. To compute (\ref{eq:cflow}), we split the dimensions $\text{z}^{D}$ of the latent variable into halfs, $\text{z}^{L} = \left\{ 1, \cdots, \nicefrac{D}{2} \right\}$ and $\text{z}^{R} = \left\{ \nicefrac{D}{2}, \cdots, d \right\}$ at each invertible layer $f_{i}$. Our transformation takes the following form for each dimension $\text{z}^{j}$ alternatively from $\text{z}^{L}$ or $\text{z}^{R}$,
\begin{align}\label{eq:cnlsq}
f_{i}^{-1}(\text{z}^{j} | \text{z}^{R}, \text{x}) = \epsilon^{j} = a(\text{z}^{R}, \text{x}) + b(\text{z}^{R}, \text{x}) \times \text{z}^{j} + \frac{c(\text{z}^{R}, \text{x})}{1 + (d(\text{z}^{R}, \text{x}) \times \text{z}^{j} + g\big(\text{z}^{R}, \text{x})\big)^2}.
\end{align}
where, $\text{z}^{j} \in \text{z}^{L}$. Details of the forward (generating) operation $f_{i}$ are in Appendix A. To ensure that the generated prior distribution is conditioned on $\text{x}$, in (\ref{eq:cnlsq}) and in the corresponding forward operation $f_{i}$, the coefficients $\left\{a,b,c,d,g\right\} \in \mathbb{R}$ are functions of both the other half of the dimensions of $\text{z}$ \emph{and} the condition $\text{x}$ (unlike \cite{ziegler2019latent}). Finally, due to the expressive power of our conditional non-linear normalizing flows, simple spherical Gaussians base distributions were sufficient.
	

\subsection{Variational Inference using Conditional Normalizing Flows based Priors}
Here, we derive the ELBO (\ref{eq:cvae}) for our novel regularized CF-VAE with our conditional flow based prior. In case of the standard CVAE with the Gaussian prior, the KL divergence term in the ELBO has a simple closed form expression. In case of our conditional flow based prior, we can use the change of variables formula to compute the KL divergence. In detail, given the base density $p(\epsilon|\text{x})$ and the Jacobian $J_{i}$ of each layer $i$ of the transformation, the log-likelihood of the latent variable $\text{z}$ under the prior can be expressed using the change of variables formula,
\begin{align}\label{eq:cflow}
\log(p_{\psi}(\text{z}|\text{x})) = \log(p(\epsilon|\text{x})) + \sum\limits_{i=1}^{n} \log( \lvert \det J_{i} \lvert ).
\end{align}

This change of variables allows us to evaluate the likelihood of latent variable $\text{z}$ over the base distribution instead of the complex conditional prior and to express the KL divergence as,
\begin{align}\label{eq:cfkl}
\begin{split}
- D_{\text{KL}}( q_{\phi}(\text{z}|\text{x},\text{y}) || p_{\psi}(\text{z}|\text{x})) &= - \mathbb{E}_{q_{\phi}(\text{z}|\text{x},\text{y})} \log(q_{\phi}(\text{z}|\text{x},\text{y})) + \mathbb{E}_{q_{\phi}(\text{z}|\text{x},\text{y})} \log(p_{\psi}(\text{z}|\text{x}))\\
&= \mathcal{H}(q_{\phi}) + \mathbb{E}_{q_{\phi}(\text{z}|\text{x},\text{y})}\log(p(\epsilon|\text{x})) + \sum\limits_{i=1}^{n} \log( \lvert \det J_{i} \lvert) .
\end{split}
\end{align}

where, $\mathcal{H}(q_{\phi})$ is the entropy of the variational distribution. Therefore, the ELBO can be expressed as,
\begin{align}\label{eq:cfcvae}
\log(p_{\theta}(\text{y}|\text{x})) \geq \mathbb{E}_{q_{\phi}(\text{z}|\text{x},\text{y})} \log(p_{\theta}(\text{y}|\text{z},\text{x})) + \mathcal{H}(q_{\phi}) + \mathbb{E}_{q_{\phi}(\text{z}|\text{x},\text{y})}\log(p(\epsilon|\text{x})) + \sum\limits_{i=1}^{n} \log( \lvert \det J_{i} \lvert)
\end{align}

\begin{wrapfigure}[13]{r}{0.4\textwidth}
  \begin{center}
    \includegraphics[width=\textwidth]{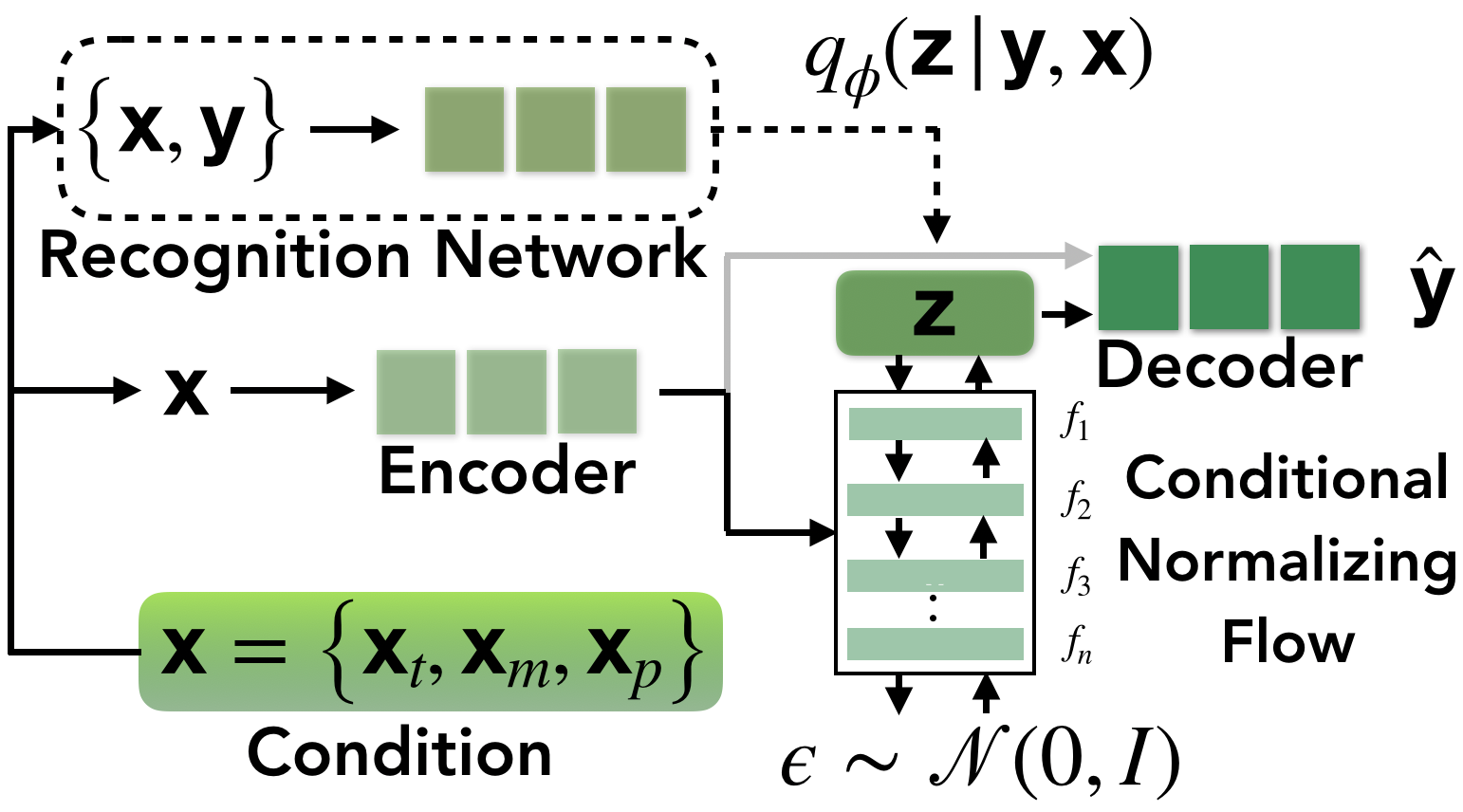}
  \end{center}
  \vspace{-0.5cm}
  \caption{CF-VAE. The decoder is regularized by removing conditioning (grey arrow) to prevent posterior collapse.}
  \label{fig:model}
\end{wrapfigure}

To learn complex conditional priors, we alternately optimize both the variational posterior distribution $q_{\phi}(\text{z}|\text{x},\text{y})$ and the conditional prior $p_{\psi}(\text{z}|\text{x})$ in (\ref{eq:cfcvae}). This would allow the variational posterior $q_{\theta}$ to match the conditional prior and vice-versa so that the ELBO (\ref{eq:cfcvae}) is maximized. However, in practice we observe instabilities during training and posterior collapse. Next, we introduce our novel regularization schemes to deal with both these problems.

\myparagraph{Posterior Regularization for Stability (pR).} The entropy and the $\log$-Jacobian of the CF-VAE objective (\ref{eq:cfcvae}) are at odds with each other. The $\log$-Jacobian favours the contraction of the base density. Therefore, $\log$-Jacobian at the right of (\ref{eq:cfcvae}) is maximized when the conditional flow maps the base distribution ($\epsilon \leftrightarrow \text{z}$ in \autoref{fig:model}) to a low entropy conditional prior and thus a low entropy variational distribution $q_{\phi}(\text{z}|\text{x},\text{y})$. Therefore, in practice we observe instabilities during training. We observe that either the entropy or the $\log$-Jacobian term dominates and the data log-likelihood is fully or partially ignored. Therefore, we regularize the posterior $q_{\phi}(\text{z}|\text{x},\text{y})$ by fixing the variance to $\text{C}$. This leads to a constant entropy term which in turn bounds the maximum possible amount of contraction, thus upper bounding the $\log$-Jacobian. This encourages our model to concentrate on explaining the data and leads better fit to the target data distribution. Note that, although $q_{\phi}(\text{z}|\text{x},\text{y})$ has fixed variance, this does not significantly effect expressivity as the marginal $q_{\phi}(\text{z}|\text{x})$ can be arbitrarily complex due to our conditional flow prior. Moreover, we observe that the LSTM based decoders employed demonstrate robust performance across a wide range of values  $\text{C}=\left[ 0.05, 0.25\right]$.
		
\myparagraph{Condition Regularization for Posterior Collapse (cR).} We observe missing modes when the target conditional data distribution has a major mode(s) and one or more minor modes (corresponding to rare events). This is because the condition $\text{x}$ on the decoder is already enough to model the main mode(s). If the cost of ignoring the minor modes is out-weighed by the cost of encoding a more complex latent distribution reflecting all modes, the minor modes and the latent variables are ignored. We propose a novel regularization scheme by removing the additional conditioning $\text{x}$ on the decoder, when the dataset in question has a dominating mode(s). This enabled by our novel conditional flow prior, which already ensures that conditioning information can be encoded in the latent space. This assumes a simpler factorization of the conditional distribution $p_{\theta}(\text{y}|\text{x}) = p_{\theta}(\text{y}|\text{z}) p_{\psi}(\text{z}|\text{x})$. This ensures that the latent variable $\text{z}$ cannot be ignored by the CF-VAE and thus must encode useful information. Note that this regularization scheme is only possible due to our conditional prior, the unconditional Gaussian prior of CVAE would always need to condition the decoder. 
\mario{i'm confused about $d\mathcal{R}$) and $p\mathcal{R}$; why is this introduced . does it have a special meaning? it does not appear in the formula. P.S.: I know understand that this is meant as a short hand and not a mathematical notation that is introduced. I suggest to also make it bold. I think part of the confusion comes from the fact that the paragraph heading is bold and the short hand is not. on a similar line I might just use regular letters ... because it is a short hand and not actual math notation. }

Finally, we discuss the integration of diverse sources of contextual information into the conditional prior $p_{\psi}(\text{z}|\text{x})$ for even richer conditional latent distributions of our regularized CF-VAE.


\subsection{Conditioning Priors on Contextual Information}
\label{sec:context}
For prediction tasks, it is often crucial to integrate sources of contextual information e.g. past trajectories or environmental information for accurate predictions. As these sources are heterogeneous, we employ source specific networks to extract fixed length vectors from each source.
	
\myparagraph{Past Trajectory.} We encode the past trajectories using a LSTM to an fixed length vector $\text{x}_{t}$. For efficiency we share the condition encoder between the conditional flow and the CF-VAE decoder.

\myparagraph{Environmental Map.} We use a CNN to encode environmental information to a set of region specific feature vectors. We apply attention conditioned on the past trajectory to extract a fixed length conditioning vector $\text{x}_{m}$, such that $\text{x}_{m}$ contains information relevant to the future trajectory.

\myparagraph{Interacting Agents.} To encode information of interacting traffic participants/agents, we build on \cite{deo2018convolutional} and propose a fully convolutional social pooling layer. We aggregate information of interacting agents using a grid overlayed on the environment. This grid is represented using a tensor, where the past trajectory information of traffic participants are aggregated into the tensor indexed corresponding to the grid in the environment. In \cite{deo2018convolutional} past trajectory information is aggregated using a LSTM. We aggregate the past trajectory information into the tensor using $1\times 1$ convolutions as it allows for stable learning and is computationally efficient. Finally, we apply several layers of $k\times k$ convolutions to capture interaction aware contextual features $\text{x}_{p}$ of traffic participants in the scene.

Due to the expressive power of our conditional non-linear normalizing flows, simple concatenation into a single vector $\text{x} = \left\{ \text{x}_{t}, \text{x}_{m}, \text{x}_{t} \right\}$ was sufficient to learn powerful conditional priors.
 

\section{Experiments}
\label{sec:experiments}
We evaluate our CF-VAE on three popular and highly multi-modal sequence prediction datasets. We begin with a description of our evaluation metrics and model architecture.

\myparagraph{Evaluation Metrics.} In line with prior work \citep{lee2017desire,bhattacharyya2018accurate,pajouheshgar2018back,deo2019scene,bhattacharyya2018bayesian}, we use the negative conditional $\log$-likelihood (-CLL) and mean Euclidean distances of the oracle Top $n$\% of $N$ predictions. The oracle Top $n$\% metric measures not only the coverage of all modes but also discourages random guessing for a reasonably large value of $n$ (e.g.\ $n=10$\%). This is because, a model can only improve this metric by moving randomly guessed samples from an overestimated mode to the correct modes (detailed analysis in Appendix F). 

\myparagraph{Conditional Flow Model Architecture.} Our conditional flow prior consists of 16 layers of conditional non-linear flows with split coupling. Increasing the number of conditional non-linear flows generally led to ``over-fitting'' on the training latent distribution.

\begin{figure*}[t]
\centering
	\resizebox{\textwidth}{!}{\begin{tabular}{ c|c@{\hskip 0.07cm}c@{\hskip 0.07cm}c|c@{\hskip 0.07cm}c@{\hskip 0.07cm}c|c }
	\toprule
    \textbf{Condition} & \multicolumn{3}{c}{\textbf{BMS-CVAE Modes \citep{bhattacharyya2018accurate}}} & \multicolumn{3}{c}{\textbf{Our CF-VAE Modes}} & \textbf{Our CF-VAE Prior} \\
	\midrule
	
	
	\includegraphics[height=2.35cm,width=2.35cm,trim={2cm 0.8cm 1cm 1.5cm},clip]{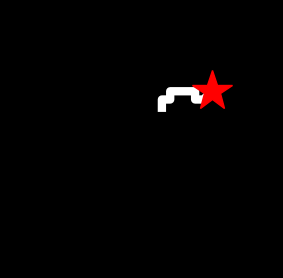} &
	\includegraphics[height=2.35cm,width=2.35cm,trim={2cm 0.8cm 1cm 1.5cm},clip]{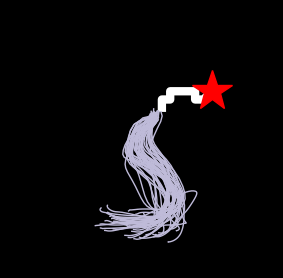} &
	\includegraphics[height=2.35cm,width=2.35cm,trim={2cm 0.8cm 1cm 1.5cm},clip]{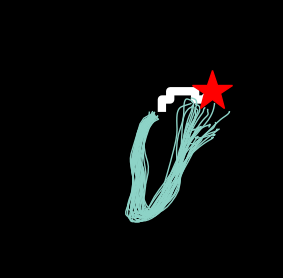} &
	\includegraphics[height=2.35cm,width=2.35cm,trim={2cm 0.8cm 1cm 1.5cm},clip]{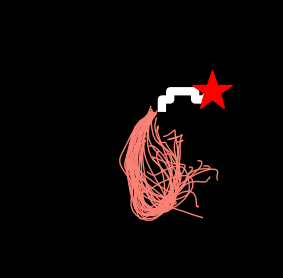} &
	\includegraphics[height=2.35cm,width=2.35cm,trim={2cm 0.8cm 1cm 1.5cm},clip]{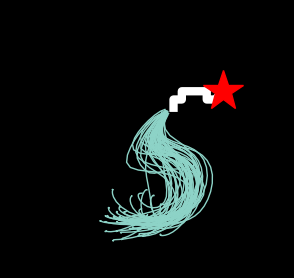} &
	\includegraphics[height=2.35cm,width=2.35cm,trim={2cm 0.8cm 1cm 1.5cm},clip]{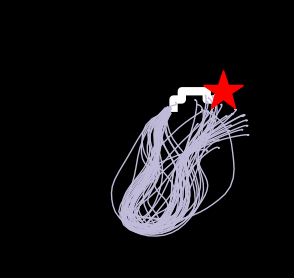} &
	\includegraphics[height=2.35cm,width=2.35cm,trim={2cm 0.8cm 1cm 1.5cm},clip]{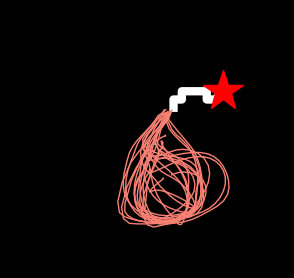} &
	\includegraphics[height=2.35cm,width=2.35cm,trim={1.5cm 1cm 1cm 0.5cm},clip]{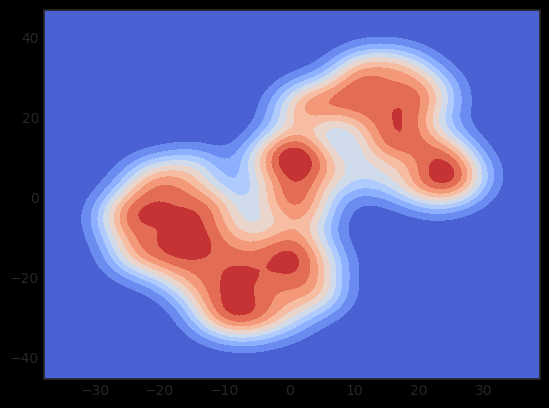} \\
	
	\includegraphics[height=2.35cm,width=2.35cm,trim={2cm 0.8cm 0.8cm 1.5cm},clip]{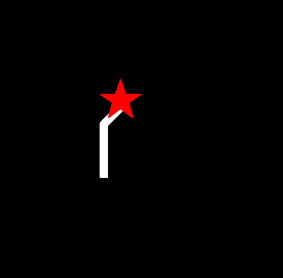} &
	\includegraphics[height=2.35cm,width=2.35cm,trim={2cm 0.8cm 0.8cm 1.5cm},clip]{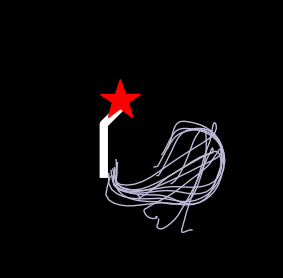} &
	\includegraphics[height=2.35cm,width=2.35cm,trim={2cm 0.8cm 0.8cm 1.5cm},clip]{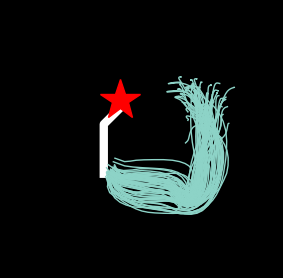} &
	\includegraphics[height=2.35cm,width=2.35cm,trim={2cm 0.8cm 0.8cm 1.5cm},clip]{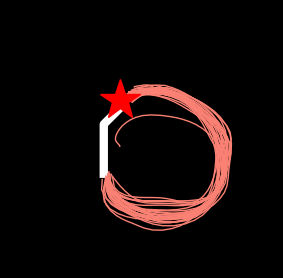} &
	\includegraphics[height=2.35cm,width=2.35cm,trim={2cm 0.8cm 0.8cm 1.5cm},clip]{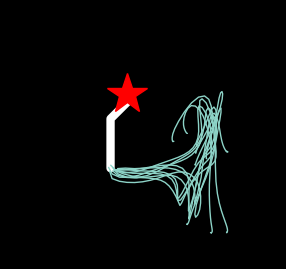} &
	\includegraphics[height=2.35cm,width=2.35cm,trim={2cm 0.8cm 0.8cm 1.5cm},clip]{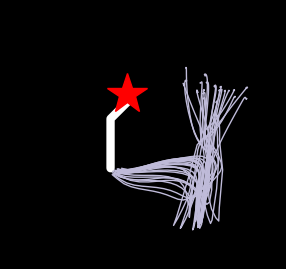} &
	\includegraphics[height=2.35cm,width=2.35cm,trim={2cm 0.8cm 0.8cm 1.5cm},clip]{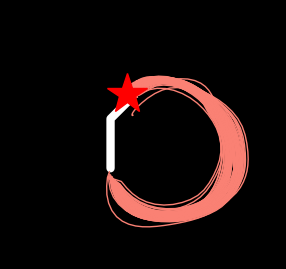} &
	\includegraphics[height=2.35cm,width=2.35cm,trim={1.5cm 1cm 0.8cm 0.5cm},clip]{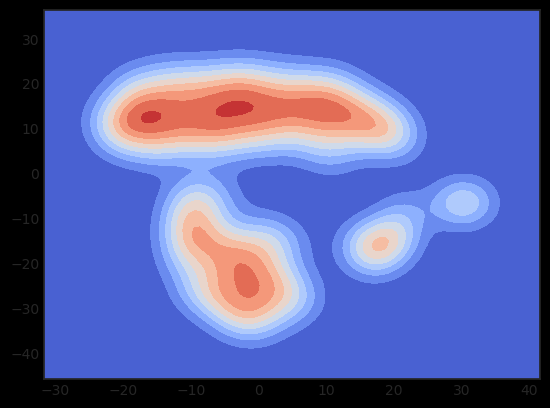} \\
	
	\includegraphics[height=2.35cm,width=2.35cm,trim={2cm 0.5cm 0.8cm 2cm},clip]{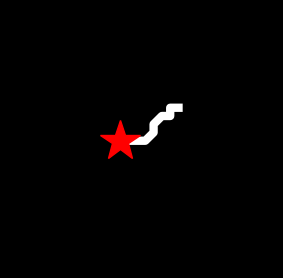} &
	\includegraphics[height=2.35cm,width=2.35cm,trim={2cm 0.5cm 0.8cm 2cm},clip]{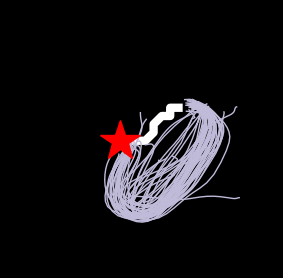} &
	\includegraphics[height=2.35cm,width=2.35cm,trim={2cm 0.5cm 0.8cm 2cm},clip]{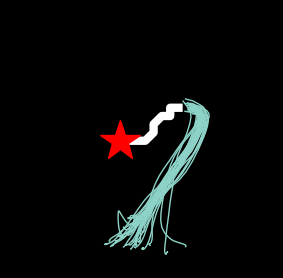} &
	\includegraphics[height=2.35cm,width=2.35cm,trim={2cm 0.5cm 0.8cm 2cm},clip]{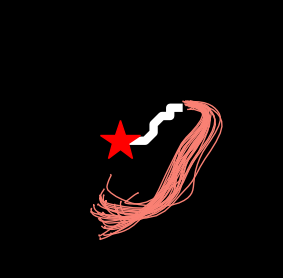} &
	\includegraphics[height=2.35cm,width=2.35cm,trim={2cm 0.5cm 0.8cm 2cm},clip]{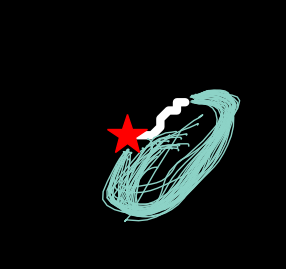} &
	\includegraphics[height=2.35cm,width=2.35cm,trim={2cm 0.5cm 0.8cm 2cm},clip]{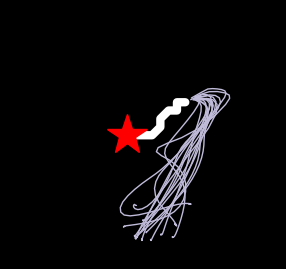} &
	\includegraphics[height=2.35cm,width=2.35cm,trim={2cm 0.5cm 0.8cm 2cm},clip]{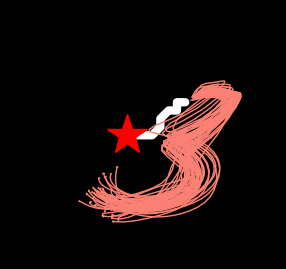} &
	\includegraphics[height=2.35cm,width=2.35cm,trim={1.5cm 1cm 0.8cm 0.5cm},clip]{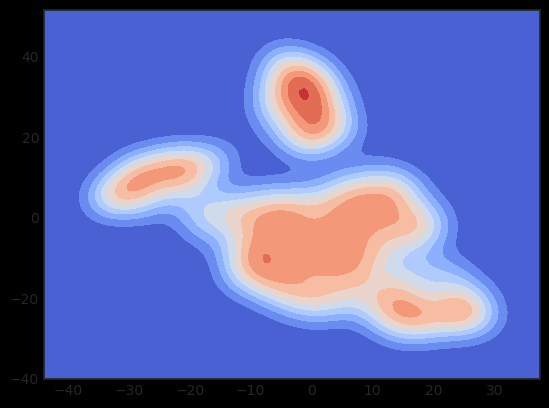} \\
	
	\bottomrule
	
	\end{tabular}}
	\vspace{-0.3cm}
\caption{Random samples clustered using k-means. The number of clusters is set manually to the number of expected digits. The corresponding priors of our CF-VAE \textbf{+} pR on the right. Note, our 64D CF-VAE latent distribution is (approximately) projected to 2D using tSNE and KDE.}
\label{fig:mnist_ex}
\end{figure*}

\subsection{MNIST Sequences} The MNIST Sequence dataset \citep{mnist_seq} consists of sequences of handwriting strokes of the MNIST digits. The state-of-the-art approach is the ``Best-of-Many''-CVAE \citep{bhattacharyya2018accurate} with a Gaussian prior. We follow the evaluation protocol of \cite{bhattacharyya2018accurate} and predict the complete stroke given the first ten steps. We also compare with, \begin{enumerate*} \item A standard CVAE with uni-modal Gaussian prior; \item A CVAE with a data dependent conditional mixture of Gaussians (MoG) prior; \item A CF-VAE without any regularization ; \item A CF-VAE without the conditional non-linear flow layers (CF-VAE\textbf{-}\emph{Affine}, replaced with affine flows \citep{lu2019structured,atanov2019semi}). \end{enumerate*} We also experiment with a conditional MoG prior (see Appendix D and E). We use the same model architecture \citep{bhattacharyya2018accurate} across all baselines.

\begin{wraptable}[12]{r}{6cm}
\centering
    \caption{Evaluation on MNIST Sequences.}
	\resizebox{\linewidth}{!}{\begin{tabular}{lc}
	\toprule
	Method & -CLL $\downarrow$ \\
	\midrule
	CVAE \citep{sohn2015learning} & 96.4\\
	BMS-CVAE \citep{bhattacharyya2018accurate} & 95.6 \\
	\midrule 
	CVAE \textbf{+} \emph{increased capacity} (Ours) & 94.5 \\ 
	\midrule
	CVAE \textbf{+} \emph{conditional prior} (Ours) & 88.9\\
	MoG-CVAE, $M=3$ & 84.6 \\
	CF-VAE \textbf{-} \emph{Affine} (Ours) & 77.2\\
	CF-VAE \textbf{-} \emph{no regularization} (Ours) & 104.3\\
	CF-VAE \textbf{+} pR, $\text{C}=0.2$ (Ours) & \textbf{74.9}\\
	\bottomrule
	\end{tabular}}
\label{tab:mnistseq}
\end{wraptable}  

We report the results in \autoref{tab:mnistseq}. We see that our CF-VAE with posterior regularization (pR) performs best. It has a performance advantage of over 20\% against the state of the art BMS-CVAE. We see that without regularization (pR) ($\text{C}=0.2$) there is a 40\% drop in performance, highlighting the effectiveness of our novel regularization scheme. We further illustrate the modes captured and the learnt multi-modal conditional flow priors in \autoref{fig:mnist_ex}. We do not use condition regularization here (cR) as we do not observe posterior collapse. In contrast, the BMS-CVAE is unable to fully capture all modes -- its predictions are pushed to the mean due to the  strong model bias induced by the Gaussian prior.  The results improve considerably with the multi-modal MoG prior ($M=3$ components work best). We also experiment with optimizing the standard CVAE architecture. This improves performance only slightly (after increasing LSTM encoder/decoder units to 256 from 48, increasing the number of layers did not help). Moreover, our experiments with a conditional (MoG) AAE/WAE \citep{gu2018dialogwae} based baseline did not improve performance beyond the standard (MoG) CVAE, because the discriminator based KL estimate tends to be an underestimate \citep{rosca2017variational}. This illustrates that in practice it is difficult to map highly multi-modal sequences to a Gaussian prior and highlights the need of a data-dependent multi-modal priors. Our CF-VAE still significantly outperforms the MoG-CVAE as normalizing flows are better at learning complex multi-modal distributions \citep{kingma2018glow}. We also see that affine conditional flow based priors leads to a drop in performance (77.2 vs 74.9 CLL) illustrating the advantage of our non-linear conditional flows.

\subsection{Stanford Drone}

\begin{table*}[!t]
\centering
\resizebox{\textwidth}{!}{
\begin{tabular}{lcccccc}
\toprule
Method & Visual & Error $@$ 1sec & Error $@$ 2sec & Error $@$ 3sec & Error $@$ 4sec & -CLL $\downarrow$ \\
\midrule
``Shotgun'' (Top 10\%) \citep{pajouheshgar2018back} & None & 0.7 & 1.7 &  3.0 &  4.5 & 91.6 \\
DESIRE-SI-IT4 (Top 10\%) \citep{lee2017desire} & RGB & 1.2 & 2.3 & 3.4 & 5.3 & x \\
STCNN (Top 10\%) \citep{pajouheshgar2018back} & RGB & 1.2   & 2.1  & 3.3 &  4.6 & x   \\
BMS-CVAE (Top 10\%) \citep{bhattacharyya2018accurate} & RGB & 0.8  & 1.7 & 3.1 & 4.6 & 126.6  \\
\midrule
MoG-CVAE, $M=3$ (Top 10\%) & None & 0.8  & 1.7 & 2.7 & 3.9 & 86.1  \\
CF-VAE \textbf{-} \emph{no regularization} (Ours, Top 10\%) & None & 0.9 & 1.9 & 3.3 & 4.7 & 96.2\\
CF-VAE \textbf{+} pR, $\text{C}=0.2$ (Ours, Top 10\%) & None & \textbf{0.7} & \textbf{1.5} & 2.5 & 3.6 & 84.6\\
CF-VAE \textbf{+} pR, $\text{C}=0.2$ (Ours, Top 10\%) & RGB  & \textbf{0.7} & \textbf{1.5} & \textbf{2.4} & \textbf{3.5} & \textbf{84.1} \\
\bottomrule
\end{tabular}}
\caption{Five fold cross validation on the Stanford Drone dataset. Euclidean error at ($\nicefrac{1}{5}$) resolution.}
\vspace{-0.3cm}
\label{tab:stanford_drone_cross}
\end{table*}

\begin{figure*}[h]
  
  \centering
  \resizebox{\textwidth}{!}{\begin{tabular}{ c@{\hskip 0.05cm}c@{\hskip 0.2cm}c@{\hskip 0.05cm}c@{\hskip 0.2cm}c@{\hskip 0.05cm}c }
	\toprule
    {Sampled Predictions} & {Latent Prior} & {Sampled Predictions} & {Latent Prior} & {Sampled Predictions} & {Latent Prior} \\ 
	\midrule
    \includegraphics[height=2cm,width=3cm,trim={0cm 0cm 0cm 0cm},clip]{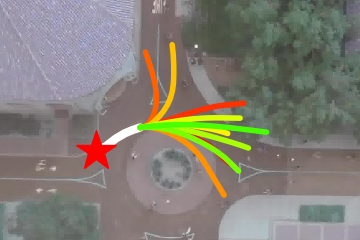} &
    \includegraphics[height=2cm,width=2cm,trim={1.5cm 1cm 0.8cm 0.5cm},clip]{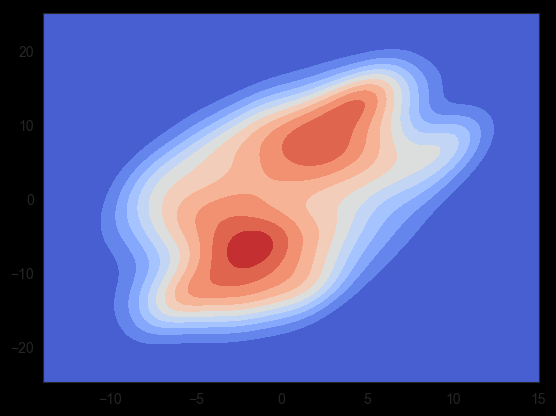} &
    \includegraphics[height=2cm,width=3cm,trim={0cm 0cm 0cm 0cm},clip]{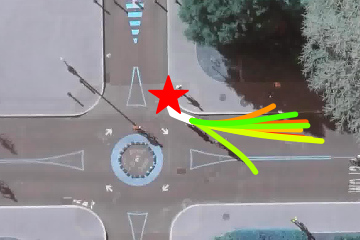} &
    \includegraphics[height=2cm,width=2cm,trim={1.5cm 1cm 0.8cm 0.5cm},clip]{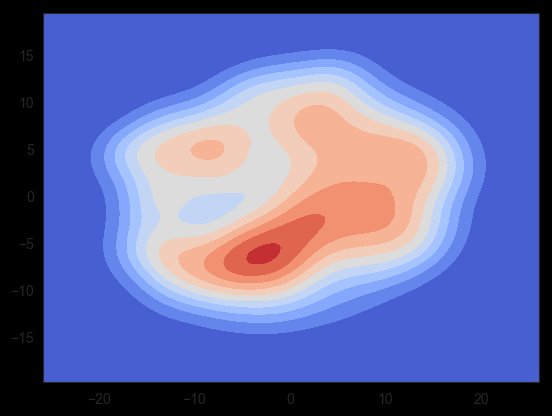} &
    \includegraphics[height=2cm,width=3cm,trim={0cm 0cm 0cm 0cm},clip]{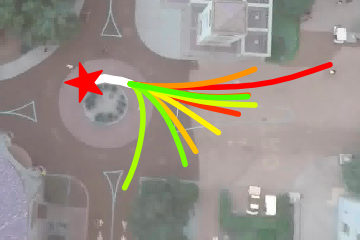} &
    \includegraphics[height=2cm,width=2cm,trim={1.5cm 1cm 0.8cm 0.5cm},clip]{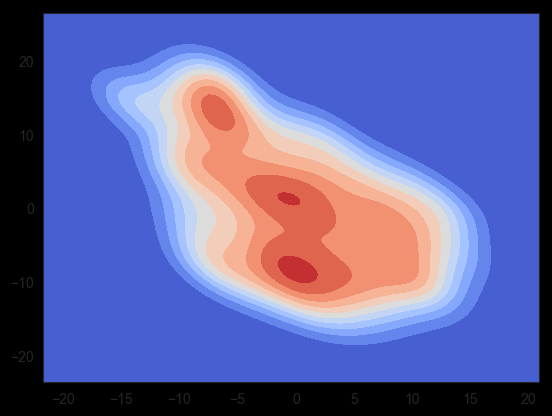} \\
	
	\includegraphics[height=2cm,width=3cm,trim={0cm 0cm 0cm 0cm},clip]{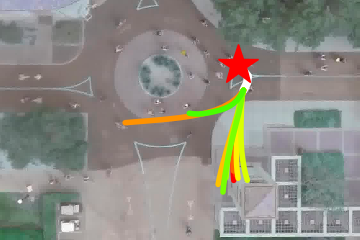} &
    \includegraphics[height=2cm,width=2cm,trim={1.5cm 1cm 0.8cm 0.5cm},clip]{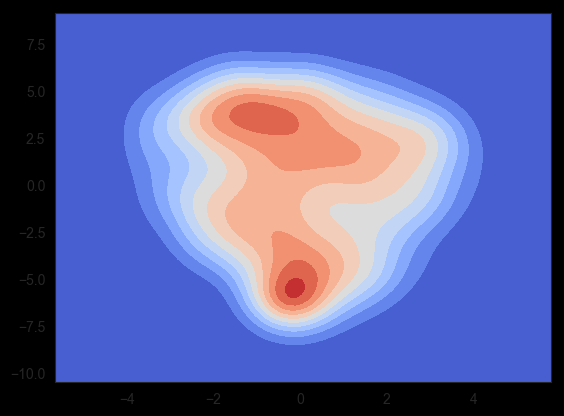} &
    \includegraphics[height=2cm,width=3cm,trim={0cm 0cm 0cm 0cm},clip]{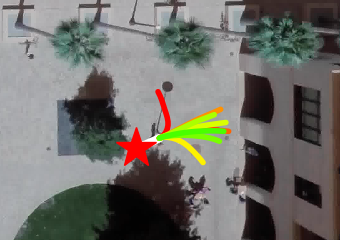} &
    \includegraphics[height=2cm,width=2cm,trim={1.5cm 1cm 0.8cm 0.5cm},clip]{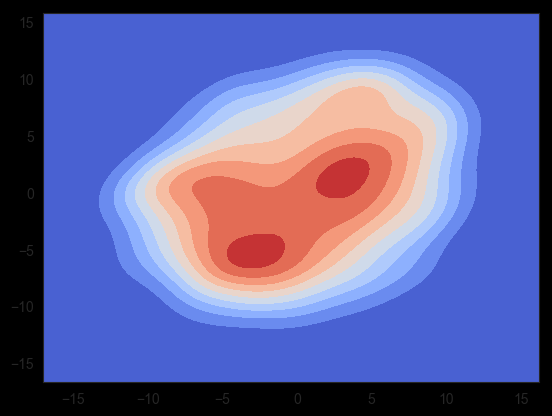} &
	\includegraphics[height=2cm,width=3cm,trim={0cm 0cm 0cm 0cm},clip]{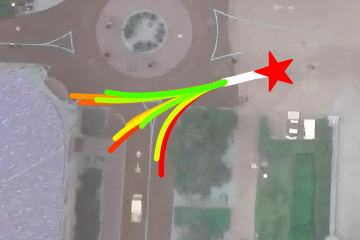} &
    \includegraphics[height=2cm,width=2cm,trim={1.5cm 1cm 0.8cm 0.5cm},clip]{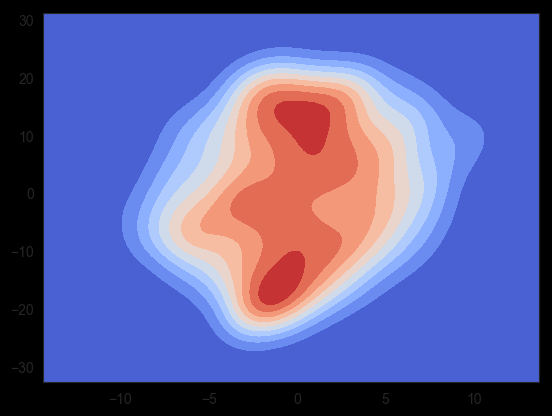} \\
	
    \bottomrule
    \end{tabular}}
    \vspace{-0.3cm}
  \caption{Randomly sampled predictions of our CF-VAE \textbf{+} pR model on the Stanford Drone. We observe that our prediction are highly multi-modal and is reflected by the Conditional Flow Priors. Note, our 64D CF-VAE latent distribution is (approximatly) projected to 2D using tSNE and KDE.}
   
  \label{fig:stanford_drone_rand_samp}
\end{figure*}

\begin{figure*}[h]
  
  \centering
  \resizebox{\textwidth}{!}{\begin{tabular}{ c@{\hskip 0.1cm}c@{\hskip 0.1cm}c@{\hskip 0.1cm}c }
    
    \includegraphics[height=2.75cm]{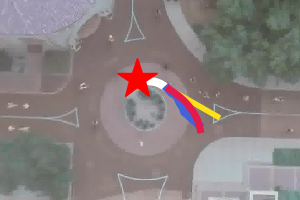} &
    \includegraphics[height=2.75cm]{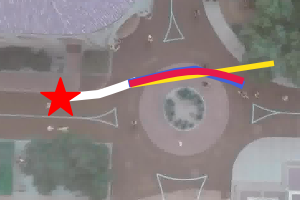} &
    \includegraphics[height=2.75cm]{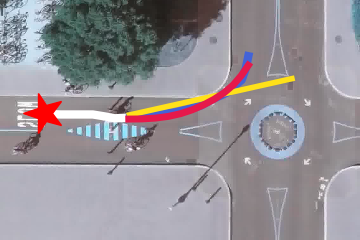} &
    \includegraphics[height=2.75cm]{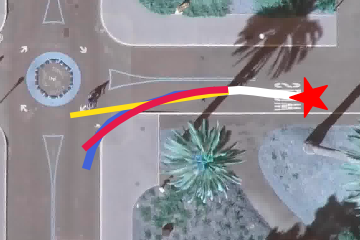} \\
    
    \end{tabular}}
    \vspace{-0.3cm}
  \caption{Comparison of our CF-VAE \textbf{+} pR (Red) and the ``Shoutgun'' baseline (Yellow) of \citep{pajouheshgar2018back}, Groundtruth (Blue). Initial conditioning trajectory in white. Our CF-VAE not only learns to capture the correct modes but also generates more fine-grained predictions.}
   
  \label{fig:stanford_drone_comp}
\end{figure*}

The Stanford Drone dataset \citep{robicquet2016learning} consists of multi-model trajectories of traffic participant e.g. pedestrians, bicyclists, cars captured from a drone. Prior works follow two different evaluation protocols, \begin{enumerate*} \item \citep{lee2017desire,bhattacharyya2018accurate,pajouheshgar2018back} use 5 fold cross validation, \item \citep{robicquet2016learning,sadeghian2018car,sadeghian2018sophie,deo2019scene} use a single split \end{enumerate*}. We evaluate using the first protocol in \autoref{tab:stanford_drone_cross} and the second in \autoref{tab:stanford_drone_stand}.

\begin{wraptable}[9]{r}{6cm}
\centering
	\resizebox{\textwidth}{!}{\begin{tabular}{lcc}
	\toprule
	Method & mADE $\downarrow$ & mFDE $\downarrow$ \\
	\midrule
	SocialGAN \citep{gupta2018social} & 27.2 & 41.4\\
	MATF GAN \citep{zhao2019multi} & 22.5 & 33.5 \\
	SoPhie \citep{sadeghian2018sophie} & 16.2 &  29.3\\
	Goal Prediction \citep{deo2019scene} &  15.7 & 28.1 \\
	\midrule
	CF-VAE \textbf{+} pR, $\text{C}=0.2$ (Ours) & \textbf{12.6} & \textbf{22.3} \\
	\bottomrule
	\end{tabular}}
	\vspace{-0.3cm}
\caption{Evaluation on the Stanford Drone dataset on a single split (see also \autoref{tab:stanford_drone_cross}).}
\label{tab:stanford_drone_stand}
\end{wraptable}

Additionally, \cite{pajouheshgar2018back} suggest a ``Shotgun'' baseline. This baseline extrapolates the trajectory from the last known position and orientation in 10 different ways -- 5 orientations: $(0^{\circ},\, \pm 8^{\circ},\, \pm 15^{\circ})$ and 5 velocities: None or exponentially weighted over the past with coefficients $(0, \, 0.3, \, 0.7, \, 1.0)$. This baseline obtains results at par with the state-of-the-art because it a good template which covers the most likely possible futures (modes) for traffic participant motion in this dataset. We report the results using 5 fold cross validation in \autoref{tab:stanford_drone_cross}. We additionally compare to a mixture of Gaussians prior (Appendix D). We use the same model architecture as in \cite{bhattacharyya2018accurate} and a CNN encoder with attention to extract features from the last observed RGB image (Appendix C). These visual features serve as additional conditioning ($\text{x}_{m}$) to our Conditional Flow model. We see that our CF-VAE model with RGB input and posterior regularization (pR) performs  best -- outperforming the state-of-art ``Shotgun'' and BMS-CVAE by over 20\% (Error $@$ 4sec). We see that our conditional flows are able to utilize visual scene (RGB) information to improve performance (3.5 vs 3.6 Error $@$ 4sec). We also see that the MoG-CVAE and our CF-VAE + pR outperforms the BMS-CVAE, even without visual scene information. This again reinforces our claim that the standard Gaussian prior induces a strong model bias and data dependent multi-modal  priors are needed for best performance. The performance advantage of CF-VAE over the MoG-CVAE again illustrates the advantage of normalizing flows at learning complex conditional multi-modal distributions. The performance advantage over the ``Shotgun'' baseline shows that our CF-VAE + pR not only learns to capture the correct modes but also generates more fine-grained predictions. The qualitative examples in \autoref{fig:stanford_drone_comp} shows that our CF-VAE is better able to capture complex trajectories with sharp turns.

We report results using the single train/test split of \citep{robicquet2016learning,sadeghian2018car,sadeghian2018sophie,deo2019scene} in \autoref{tab:stanford_drone_stand}. We use the minimum Average Displacement Error (mADE) and minimum Final Displacement Error (mFDE) metrics as in \citep{deo2019scene}. The minimum is over as set of predictions of size $N$. Although this metric is less robust to random guessing compared to the Top $n$\% metric, it avoids rewarding random guessing for a small enough value of $N$. We choose $N=20$ as in \citep{deo2019scene}. Similar to the results with 5 fold cross validation, we observe 20\% improvement over the state-of-the-art.

\subsection{HighD}
The HighD dataset \citep{highDdataset} consists of vehicle trajectories recorded using a drone over highways. In contrast to other vehicle trajectory datasets e.g. NGSIM it contains minimal false positive trajectory collisions or physically improvable velocities.

\begin{wraptable}[10]{r}{7.5cm}
\centering
	\resizebox{\textwidth}{!}{\begin{tabular}{lcccc}
	\toprule
	Method & Context & ADE $\downarrow$ & FDE $\downarrow$ & -CLL $\downarrow$ \\
	\midrule
	Constant Velocity & None & 1.09 & 2.66 & x\\
	FF \citep{diehl2019graph} & None & 0.45 & 1.09 & x\\
	GAT \citep{diehl2019graph}& Yes &  0.47 & 1.04 & x\\
	CVAE (Top 10\%) & None &  0.45 & 0.96 & 5.32 \\
	CVAE \textbf{+} \emph{Cyclic KL} (Top 10\%) & None &  0.38 & 0.80 & 4.80 \\
	\midrule
	CF-VAE \textbf{+} pR, (Ours, Top 10\%) & None & 0.44 & 0.94 & 4.71\\
	CF-VAE \textbf{+} $\left\{\text{pR,cR}\right\}$, (Ours, Top 10\%) & None & 0.30 & 0.57 & 3.64\\
	CF-VAE \textbf{+} $\left\{\text{pR,cR}\right\}$, (Ours, Top 10\%) & Yes  & \textbf{0.29} & \textbf{0.55} & \textbf{3.42}	\\
	\bottomrule
	\end{tabular}}
	\vspace{-0.3cm}
\caption{Evaluation on the HighD dataset.}
\label{tab:highd_eval}
\end{wraptable} 

The HighD dataset is challenging because lane changes or interactions are rare $\sim$  10\% of all trajectories. The distribution of future trajectories contain a single main mode (linear continuations) along with several minor modes. Thus, approaches which predict a single mean trajectory (targeting the main mode) are challenging to outperform. In \autoref{tab:highd_eval}, we see that the simple Feed Forward (FF) model performs well and the Graph Convolutional GAT model of \cite{diehl2019graph}, which captures interactions, only narrowly outperforms the FF model. This dataset is challenging for CVAE based models as they frequently suffer from posterior collapse when a single mode dominates. This is clearly observed with our CVAE baseline in \autoref{tab:highd_eval}. To prevent posterior collapse, we use the cyclic KL annealing scheme proposed in \cite{liu2019cyclical} (using a MoG prior did not help). This already leads to significant improvement over the deterministic FF and GAT baselines. We also observe posterior collapse with our CF-VAE model. Therefore, we regularize by removing additional conditioning (cR). Our CF-VAE \textbf{+} $\left\{\text{pR,cR}\right\}$ with condition regularization significantly outperforms the CF-VAE \textbf{+} pR and CVAE baselines (with cyclic KL annealing), demonstrating the effectiveness of our condition regularization scheme (cR) in preventing posterior collapse. The addition of contextual information of interacting traffic participants using our convolutional social pooling network with 1$\times$1 convolutions significantly improves performance (also see Appendix G), demonstrating the effectiveness of our conditional normalizing flow based priors.


\section{Conclusion}
\label{sec:conclusion}
In this work, we presented the first variational model for learning multi-modal conditional data distributions with Conditional Flow based priors -- the Conditional Flow Variational Autoencoder (CF-VAE). Furthermore, we propose two novel regularization techniques -- posterior regularization (pR) and condition regularization (cR) -- which stabilizes training solutions and prevents posterior collapse leading to better fit to the target distribution. This techniques lead to better match to the target distribution. Our \marioo{not sure what "rigorous" means in this context. sounds a bit like an empty phrase.} experiments on diverse sequence prediction datasets show that our CF-VAE achieves state-of-the-art results across different performance metrics.

\marioo{right now performance is out main argument that it works. but this does not give direct support to our contributions. is there anything that we can mentioned that is closer to our contributions that would verify that we "did the right thing" or that our method works as advertised? something like the prior works better - or the improvement on multimodal/uncertain secquences is better / more improvement?}




\clearpage


\bibliography{iclr2020_conference}  

\begin{thebibliography}{54}
\providecommand{\natexlab}[1]{#1}
\providecommand{\url}[1]{\texttt{#1}}
\expandafter\ifx\csname urlstyle\endcsname\relax
  \providecommand{\doi}[1]{doi: #1}\else
  \providecommand{\doi}{doi: \begingroup \urlstyle{rm}\Url}\fi

\bibitem[Alahi et~al.(2016)Alahi, Goel, Ramanathan, Robicquet, Fei-Fei, and
  Savarese]{alahi2016social}
Alexandre Alahi, Kratarth Goel, Vignesh Ramanathan, Alexandre Robicquet,
  Li~Fei-Fei, and Silvio Savarese.
\newblock Social lstm: Human trajectory prediction in crowded spaces.
\newblock In \emph{CVPR}, 2016.

\bibitem[Ardizzone et~al.(2019)Ardizzone, Kruse, Wirkert, Rahner, Pellegrini,
  Klessen, Maier-Hein, Rother, and K{\"o}the]{ardizzone2018analyzing}
Lynton Ardizzone, Jakob Kruse, Sebastian Wirkert, Daniel Rahner, Eric~W
  Pellegrini, Ralf~S Klessen, Lena Maier-Hein, Carsten Rother, and Ullrich
  K{\"o}the.
\newblock Analyzing inverse problems with invertible neural networks.
\newblock In \emph{ICLR}, 2019.

\bibitem[Atanov et~al.(2019)Atanov, Volokhova, Ashukha, Sosnovik, and
  Vetrov]{atanov2019semi}
Andrei Atanov, Alexandra Volokhova, Arsenii Ashukha, Ivan Sosnovik, and Dmitry
  Vetrov.
\newblock Semi-conditional normalizing flows for semi-supervised learning.
\newblock In \emph{ICML Workshop}, 2019.

\bibitem[Babaeizadeh et~al.(2018)Babaeizadeh, Finn, Erhan, Campbell, and
  Levine]{babaeizadeh2017stochastic}
Mohammad Babaeizadeh, Chelsea Finn, Dumitru Erhan, Roy~H Campbell, and Sergey
  Levine.
\newblock Stochastic variational video prediction.
\newblock In \emph{ICLR}, 2018.

\bibitem[Behrmann et~al.(2019)Behrmann, Duvenaud, and
  Jacobsen]{behrmann2018invertible}
Jens Behrmann, David Duvenaud, and J{\"o}rn-Henrik Jacobsen.
\newblock Invertible residual networks.
\newblock In \emph{ICML}, 2019.

\bibitem[Berg et~al.(2018)Berg, Hasenclever, Tomczak, and
  Welling]{berg2018sylvester}
Rianne van~den Berg, Leonard Hasenclever, Jakub~M Tomczak, and Max Welling.
\newblock Sylvester normalizing flows for variational inference.
\newblock In \emph{UAI}, 2018.

\bibitem[Bhattacharyya et~al.(2018)Bhattacharyya, Schiele, and
  Fritz]{bhattacharyya2018accurate}
Apratim Bhattacharyya, Bernt Schiele, and Mario Fritz.
\newblock Accurate and diverse sampling of sequences based on a “best of
  many” sample objective.
\newblock In \emph{CVPR}, 2018.

\bibitem[Bhattacharyya et~al.(2019)Bhattacharyya, Fritz, and
  Schiele]{bhattacharyya2018bayesian}
Apratim Bhattacharyya, Mario Fritz, and Bernt Schiele.
\newblock Bayesian prediction of future street scenes using synthetic
  likelihoods.
\newblock In \emph{ICLR}, 2019.

\bibitem[Bowman et~al.(2016)Bowman, Vilnis, Vinyals, Dai, Jozefowicz, and
  Bengio]{bowman2015generating}
Samuel~R Bowman, Luke Vilnis, Oriol Vinyals, Andrew~M Dai, Rafal Jozefowicz,
  and Samy Bengio.
\newblock Generating sentences from a continuous space.
\newblock \emph{CONLL}, 2016.

\bibitem[Chen et~al.(2017)Chen, Kingma, Salimans, Duan, Dhariwal, Schulman,
  Sutskever, and Abbeel]{chen2016variational}
Xi~Chen, Diederik~P Kingma, Tim Salimans, Yan Duan, Prafulla Dhariwal, John
  Schulman, Ilya Sutskever, and Pieter Abbeel.
\newblock Variational lossy autoencoder.
\newblock In \emph{ICLR}, 2017.

\bibitem[D.~De~Jong(2016)]{mnist_seq}
Edwin D.~De~Jong.
\newblock The mnist sequence dataset.
\newblock \url{https://edwin-de-jong.github.io/blog/mnist-sequence-data/},
  2016.
\newblock Accessed: 2019-07-07.

\bibitem[Deo \& Trivedi(2018)Deo and Trivedi]{deo2018convolutional}
Nachiket Deo and Mohan~M Trivedi.
\newblock Convolutional social pooling for vehicle trajectory prediction.
\newblock In \emph{CVPR Workshop}, 2018.

\bibitem[Deo \& Trivedi(2019)Deo and Trivedi]{deo2019scene}
Nachiket Deo and Mohan~M Trivedi.
\newblock Scene induced multi-modal trajectory forecasting via planning.
\newblock In \emph{ICRA Workshop}, 2019.

\bibitem[Diehl et~al.(2019)Diehl, Brunner, Le, and Knoll]{diehl2019graph}
Frederik Diehl, Thomas Brunner, Michael~Truong Le, and Alois Knoll.
\newblock Graph neural networks for modelling traffic participant interaction.
\newblock In \emph{ITSC}, 2019.

\bibitem[Dieng et~al.(2019)Dieng, Kim, Rush, and Blei]{dieng2018avoiding}
Adji~B Dieng, Yoon Kim, Alexander~M Rush, and David~M Blei.
\newblock Avoiding latent variable collapse with generative skip models.
\newblock \emph{AISTATS}, 2019.

\bibitem[Dinh et~al.(2015)Dinh, Krueger, and Bengio]{dinh2014nice}
Laurent Dinh, David Krueger, and Yoshua Bengio.
\newblock Nice: Non-linear independent components estimation.
\newblock In \emph{ICLR}, 2015.

\bibitem[Dinh et~al.(2017)Dinh, Sohl-Dickstein, and Bengio]{dinh2016density}
Laurent Dinh, Jascha Sohl-Dickstein, and Samy Bengio.
\newblock Density estimation using real nvp.
\newblock In \emph{ICLR}, 2017.

\bibitem[Goyal et~al.(2017)Goyal, Hu, Liang, Wang, and
  Xing]{goyal2017nonparametric}
Prasoon Goyal, Zhiting Hu, Xiaodan Liang, Chenyu Wang, and Eric~P Xing.
\newblock Nonparametric variational auto-encoders for hierarchical
  representation learning.
\newblock In \emph{ICCV}, 2017.

\bibitem[Gu et~al.(2018)Gu, Cho, Ha, and Kim]{gu2018dialogwae}
Xiaodong Gu, Kyunghyun Cho, Jung-Woo Ha, and Sunghun Kim.
\newblock Dialogwae: Multimodal response generation with conditional
  wasserstein auto-encoder.
\newblock \emph{arXiv preprint arXiv:1805.12352}, 2018.

\bibitem[Gulrajani et~al.(2017)Gulrajani, Kumar, Ahmed, Taiga, Visin, Vazquez,
  and Courville]{gulrajani2016pixelvae}
Ishaan Gulrajani, Kundan Kumar, Faruk Ahmed, Adrien~Ali Taiga, Francesco Visin,
  David Vazquez, and Aaron Courville.
\newblock Pixelvae: A latent variable model for natural images.
\newblock In \emph{ICLR}, 2017.

\bibitem[Gupta et~al.(2018)Gupta, Johnson, Fei-Fei, Savarese, and
  Alahi]{gupta2018social}
Agrim Gupta, Justin Johnson, Li~Fei-Fei, Silvio Savarese, and Alexandre Alahi.
\newblock Social gan: Socially acceptable trajectories with generative
  adversarial networks.
\newblock In \emph{CVPR}, 2018.

\bibitem[Helbing \& Molnar(1995)Helbing and Molnar]{helbing1995social}
Dirk Helbing and Peter Molnar.
\newblock Social force model for pedestrian dynamics.
\newblock \emph{Physical review E}, 51, 1995.

\bibitem[Higgins et~al.(2017)Higgins, Matthey, Pal, Burgess, Glorot, Botvinick,
  Mohamed, and Lerchner]{higgins2017beta}
Irina Higgins, Loic Matthey, Arka Pal, Christopher Burgess, Xavier Glorot,
  Matthew Botvinick, Shakir Mohamed, and Alexander Lerchner.
\newblock beta-vae: Learning basic visual concepts with a constrained
  variational framework.
\newblock In \emph{ICLR}, 2017.

\bibitem[Hoffman \& Johnson(2016)Hoffman and Johnson]{hoffman2016elbo}
Matthew~D Hoffman and Matthew~J Johnson.
\newblock Elbo surgery: yet another way to carve up the variational evidence
  lower bound.
\newblock In \emph{NIPS Workshop}, 2016.

\bibitem[Holmes()]{holmes2002use}
GC~Holmes.
\newblock The use of hyperbolic cosines in solving cubic polynomials.
\newblock \emph{The Mathematical Gazette}.

\bibitem[Huang et~al.(2018)Huang, Krueger, Lacoste, and
  Courville]{huang2018neural}
Chin-Wei Huang, David Krueger, Alexandre Lacoste, and Aaron Courville.
\newblock Neural autoregressive flows.
\newblock In \emph{ICML}, 2018.

\bibitem[Kingma \& Welling(2014)Kingma and Welling]{kingma2013auto}
Diederik~P Kingma and Max Welling.
\newblock Auto-encoding variational bayes.
\newblock In \emph{ICLR}, 2014.

\bibitem[Kingma \& Dhariwal(2018)Kingma and Dhariwal]{kingma2018glow}
Durk~P Kingma and Prafulla Dhariwal.
\newblock Glow: Generative flow with invertible 1x1 convolutions.
\newblock In \emph{NeurIPS}, 2018.

\bibitem[Kingma et~al.(2016)Kingma, Salimans, Jozefowicz, Chen, Sutskever, and
  Welling]{kingma2016improved}
Durk~P Kingma, Tim Salimans, Rafal Jozefowicz, Xi~Chen, Ilya Sutskever, and Max
  Welling.
\newblock Improved variational inference with inverse autoregressive flow.
\newblock In \emph{NIPS}, 2016.

\bibitem[Krajewski et~al.(2018)Krajewski, Bock, Kloeker, and
  Eckstein]{highDdataset}
Robert Krajewski, Julian Bock, Laurent Kloeker, and Lutz Eckstein.
\newblock The highd dataset: A drone dataset of naturalistic vehicle
  trajectories on german highways for validation of highly automated driving
  systems.
\newblock In \emph{ITSC}, 2018.

\bibitem[Kumar et~al.(2019)Kumar, Babaeizadeh, Erhan, Finn, Levine, Dinh, and
  Kingma]{kumar2019videoflow}
Manoj Kumar, Mohammad Babaeizadeh, Dumitru Erhan, Chelsea Finn, Sergey Levine,
  Laurent Dinh, and Durk Kingma.
\newblock Videoflow: A flow-based generative model for video.
\newblock \emph{arXiv preprint arXiv:1903.01434}, 2019.

\bibitem[Lee et~al.(2017)Lee, Choi, Vernaza, Choy, Torr, and
  Chandraker]{lee2017desire}
Namhoon Lee, Wongun Choi, Paul Vernaza, Christopher~B Choy, Philip~HS Torr, and
  Manmohan Chandraker.
\newblock Desire: Distant future prediction in dynamic scenes with interacting
  agents.
\newblock In \emph{CVPR}, 2017.

\bibitem[Liu et~al.(2019)Liu, Gao, Celikyilmaz, Carin, et~al.]{liu2019cyclical}
Xiaodong Liu, Jianfeng Gao, Asli Celikyilmaz, Lawrence Carin, et~al.
\newblock Cyclical annealing schedule: A simple approach to mitigating kl
  vanishing.
\newblock In \emph{NAACL}, 2019.

\bibitem[Lu \& Huang(2019)Lu and Huang]{lu2019structured}
You Lu and Bert Huang.
\newblock Structured output learning with conditional generative flows.
\newblock In \emph{ICML Workshop}, 2019.

\bibitem[Nalisnick \& Smyth(2017)Nalisnick and Smyth]{nalisnick2016stick}
Eric Nalisnick and Padhraic Smyth.
\newblock Stick-breaking variational autoencoders.
\newblock In \emph{ICLR}, 2017.

\bibitem[Pajouheshgar \& Lampert(2018)Pajouheshgar and
  Lampert]{pajouheshgar2018back}
Ehsan Pajouheshgar and Christoph~H Lampert.
\newblock Back to square one: probabilistic trajectory forecasting without
  bells and whistles.
\newblock In \emph{NeurIPs Workshop}, 2018.

\bibitem[Razavi et~al.(2019)Razavi, Oord, Poole, and
  Vinyals]{razavi2019preventing}
Ali Razavi, A{\"a}ron van~den Oord, Ben Poole, and Oriol Vinyals.
\newblock Preventing posterior collapse with delta-vaes.
\newblock In \emph{ICLR}, 2019.

\bibitem[Rezende \& Mohamed(2015)Rezende and Mohamed]{rezende2015variational}
Danilo~Jimenez Rezende and Shakir Mohamed.
\newblock Variational inference with normalizing flows.
\newblock In \emph{ICML}, 2015.

\bibitem[Rhinehart et~al.(2018)Rhinehart, Kitani, and
  Vernaza]{rhinehart2018r2p2}
Nicholas Rhinehart, Kris~M Kitani, and Paul Vernaza.
\newblock R2p2: A reparameterized pushforward policy for diverse, precise
  generative path forecasting.
\newblock In \emph{ECCV}, 2018.

\bibitem[Robicquet et~al.(2016)Robicquet, Sadeghian, Alahi, and
  Savarese]{robicquet2016learning}
Alexandre Robicquet, Amir Sadeghian, Alexandre Alahi, and Silvio Savarese.
\newblock Learning social etiquette: Human trajectory understanding in crowded
  scenes.
\newblock In \emph{ECCV}, 2016.

\bibitem[Rosca et~al.(2017)Rosca, Lakshminarayanan, Warde-Farley, and
  Mohamed]{rosca2017variational}
Mihaela Rosca, Balaji Lakshminarayanan, David Warde-Farley, and Shakir Mohamed.
\newblock Variational approaches for auto-encoding generative adversarial
  networks.
\newblock \emph{arXiv preprint arXiv:1706.04987}, 2017.

\bibitem[Sadeghian et~al.(2018)Sadeghian, Legros, Voisin, Vesel, Alahi, and
  Savarese]{sadeghian2018car}
Amir Sadeghian, Ferdinand Legros, Maxime Voisin, Ricky Vesel, Alexandre Alahi,
  and Silvio Savarese.
\newblock Car-net: Clairvoyant attentive recurrent network.
\newblock In \emph{ECCV}, 2018.

\bibitem[Sadeghian et~al.(2019)Sadeghian, Kosaraju, Sadeghian, Hirose,
  Rezatofighi, and Savarese]{sadeghian2018sophie}
Amir Sadeghian, Vineet Kosaraju, Ali Sadeghian, Noriaki Hirose, S~Hamid
  Rezatofighi, and Silvio Savarese.
\newblock Sophie: An attentive gan for predicting paths compliant to social and
  physical constraints.
\newblock In \emph{CVPR}, 2019.

\bibitem[Sohn et~al.(2015)Sohn, Lee, and Yan]{sohn2015learning}
Kihyuk Sohn, Honglak Lee, and Xinchen Yan.
\newblock Learning structured output representation using deep conditional
  generative models.
\newblock In \emph{NIPS}, 2015.

\bibitem[Tabak et~al.(2010)Tabak, Vanden-Eijnden, et~al.]{tabak2010density}
Esteban~G Tabak, Eric Vanden-Eijnden, et~al.
\newblock Density estimation by dual ascent of the log-likelihood.
\newblock In \emph{Communications in Mathematical Sciences}, volume~8, 2010.

\bibitem[Tolstikhin et~al.(2017)Tolstikhin, Bousquet, Gelly, and
  Schoelkopf]{tolstikhin2017wasserstein}
Ilya Tolstikhin, Olivier Bousquet, Sylvain Gelly, and Bernhard Schoelkopf.
\newblock Wasserstein auto-encoders.
\newblock \emph{arXiv preprint arXiv:1711.01558}, 2017.

\bibitem[Tomczak \& Welling(2016)Tomczak and Welling]{tomczak2016improving}
Jakub~M Tomczak and Max Welling.
\newblock Improving variational auto-encoders using householder flow.
\newblock In \emph{NIPS Workshop}, 2016.

\bibitem[Tomczak \& Welling(2018)Tomczak and Welling]{tomczak2017vae}
Jakub~M Tomczak and Max Welling.
\newblock Vae with a vampprior.
\newblock In \emph{AISTATS}, 2018.

\bibitem[Wang et~al.(2017)Wang, Schwing, and Lazebnik]{wang2017diverse}
Liwei Wang, Alexander Schwing, and Svetlana Lazebnik.
\newblock Diverse and accurate image description using a variational
  auto-encoder with an additive gaussian encoding space.
\newblock In \emph{Advances in Neural Information Processing Systems}, pp.\
  5756--5766, 2017.

\bibitem[Wang \& Wang(2019)Wang and Wang]{wang2019riemannian}
Prince~Zizhuang Wang and William~Yang Wang.
\newblock Riemannian normalizing flow on variational wasserstein autoencoder
  for text modeling.
\newblock In \emph{NAACL}, 2019.

\bibitem[Yang et~al.(2017)Yang, Hu, Salakhutdinov, and
  Berg-Kirkpatrick]{yang2017improved}
Zichao Yang, Zhiting Hu, Ruslan Salakhutdinov, and Taylor Berg-Kirkpatrick.
\newblock Improved variational autoencoders for text modeling using dilated
  convolutions.
\newblock In \emph{ICML}, 2017.

\bibitem[Zhao et~al.(2017)Zhao, Song, and Ermon]{zhao2017infovae}
Shengjia Zhao, Jiaming Song, and Stefano Ermon.
\newblock Infovae: Information maximizing variational autoencoders.
\newblock In \emph{arXiv preprint arXiv:1706.02262}, 2017.

\bibitem[Zhao et~al.(2019)Zhao, Xu, Monfort, Choi, Baker, Zhao, Wang, and
  Nian~Wu]{zhao2019multi}
Tianyang Zhao, Yifei Xu, Mathew Monfort, Wongun Choi, Chris Baker, Yibiao Zhao,
  Yizhou Wang, and Ying Nian~Wu.
\newblock Multi-agent tensor fusion for contextual trajectory prediction.
\newblock In \emph{CVPR}, 2019.

\bibitem[Ziegler \& Rush(2019)Ziegler and Rush]{ziegler2019latent}
Zachary~M Ziegler and Alexander~M Rush.
\newblock Latent normalizing flows for discrete sequences.
\newblock In \emph{ICML}, 2019.

\end{thebibliography}
\bibliographystyle{iclr2020_conference}


\appendix
\clearpage

\section*{Appendix A. Conditional Non-Linear Normalizing Flows}
In Subsection 3.1 of the main paper, we describe the inverse operation $f_{i}^{-1}$ of our non-linear conditional normalizing flows. Here, we describe the forward operation. Note that while the forward operation is necessary to compute the likelihood (3) (in the main paper) during training, the forward operation is necessary to sample from the latent prior distribution of our CF-VAE. The forward operation consists of solving for the roots of the following equation (more details in \citep{ziegler2019latent}),
\begin{align}\label{eq:cnlsq_f}
\begin{split}
& - b d^2 (\epsilon^{j})^3 + ((\text{z}^{j} - a) d^{2} - 2 d g b) (\epsilon^{j})^{2} \\
& +(2 d g (\text{z}^{j} - a) - b(g^{2} + 1)) \epsilon^{j} + ((\text{z}^{j} - a)(g^{2} + 1) - c) = 0
\end{split}
\end{align}

This equation has one real root which can be found analytically \citep{holmes2002use}. As mentioned in the main paper, note that the coefficients $\left\{a,b,c,d,g\right\}$ are also functions of the condition $\text{x}$ (unlike \citep{ziegler2019latent}). 


\section*{Appendix B. Additional Evaluation of Conditional Non-Linear Flows}

\begin{figure*}[h]
  
  \centering
  \resizebox{\textwidth}{!}{\begin{tabular}{ c@{\hskip 1.5cm}c@{\hskip 1.5cm}c@{\hskip 1.5cm}c }
  
	\toprule
	Given $\text{x}$ in, & $p(\text{y} | \text{x})$ & Cond Affine Flow & \textbf{Our Cond NL Flow} \\
	\midrule
    
    \includegraphics[height=2.5cm]{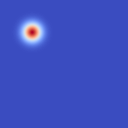} &
    \includegraphics[height=2.5cm]{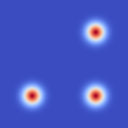} &
    \includegraphics[height=2.5cm]{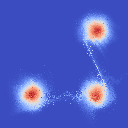} &
    \includegraphics[height=2.5cm]{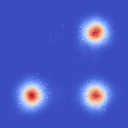} \\
	
	\includegraphics[height=2.5cm]{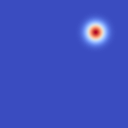} &
    \includegraphics[height=2.5cm]{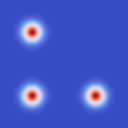} &
    \includegraphics[height=2.5cm]{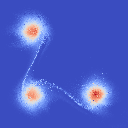} &
    \includegraphics[height=2.5cm]{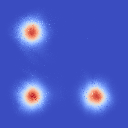} \\
	
	\includegraphics[height=2.5cm]{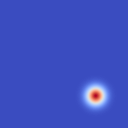} &
    \includegraphics[height=2.5cm]{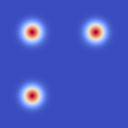} &
    \includegraphics[height=2.5cm]{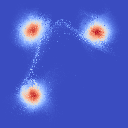} &
    \includegraphics[height=2.5cm]{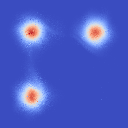} \\
	
	\includegraphics[height=2.5cm]{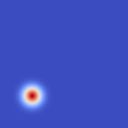} &
    \includegraphics[height=2.5cm]{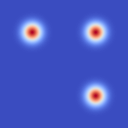} &
    \includegraphics[height=2.5cm]{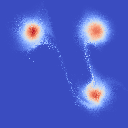} &
    \includegraphics[height=2.5cm]{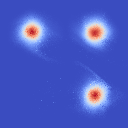} \\
	
	\bottomrule
    
    \end{tabular}}
  \caption{Comparison between conditional affine flows of \citep{atanov2019semi,lu2019structured} and our conditional non-linear (Cond NL) flows. We see that the conditional affine flows cannot fully capture multi-modal distributions (``tails'' between modes), while our conditional non-linear flows does not have distinctive ``tails''.}
   
  \label{fig:cnlsq_mog}
\end{figure*}

\begin{figure*}[h]
  
  \centering
  \resizebox{\textwidth}{!}{\begin{tabular}{ c@{\hskip 1.5cm}c@{\hskip 1.5cm}c@{\hskip 1.5cm}c }
  
	\toprule
	Given $\text{x}$ in, & $p(\text{y} | \text{x})$ & Cond Affine Flow & \textbf{Our Cond NL Flow} \\
	\midrule
    
    \includegraphics[height=2.5cm,trim={0.4cm 0.4cm 0.4cm 0.4cm},clip]{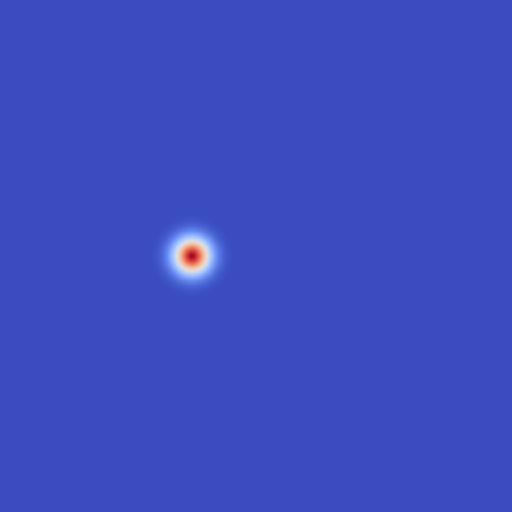} &
    \includegraphics[height=2.5cm,trim={0.4cm 0.4cm 0.4cm 0.4cm},clip]{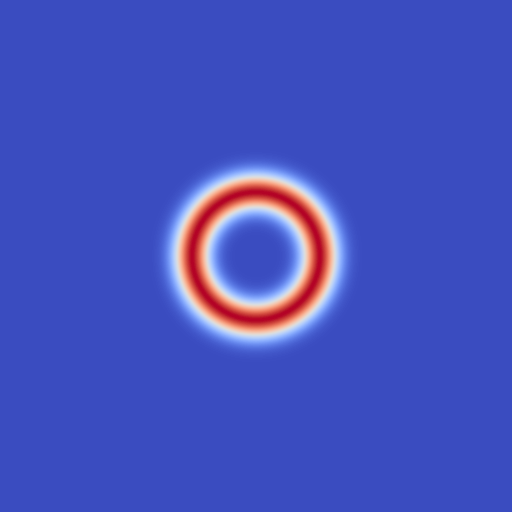} &
    \includegraphics[height=2.5cm,trim={0.4cm 0.4cm 0.4cm 0.4cm},clip]{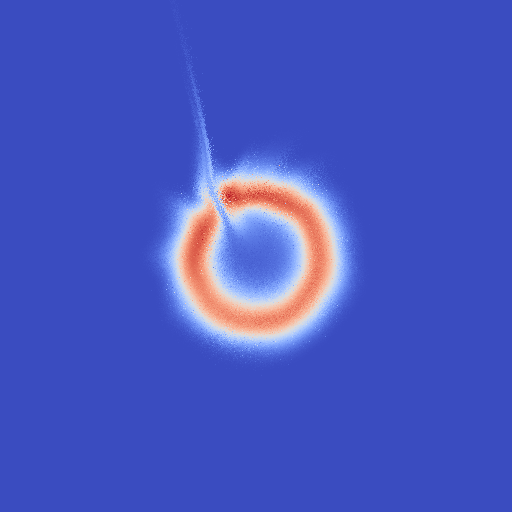} &
    \includegraphics[height=2.5cm,trim={0.4cm 0.4cm 0.4cm 0.4cm},clip]{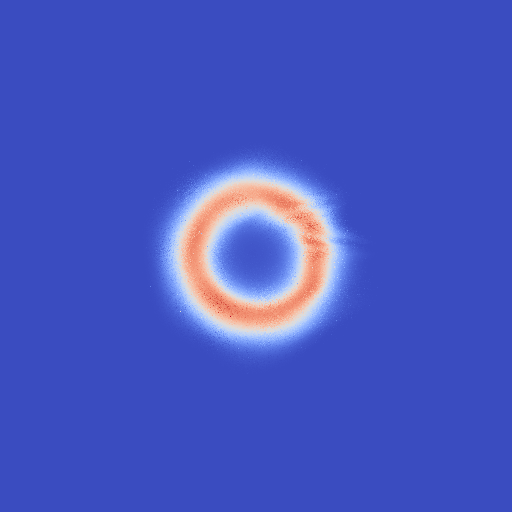} \\
	
	\includegraphics[height=2.5cm,trim={0.4cm 0.4cm 0.4cm 0.4cm},clip]{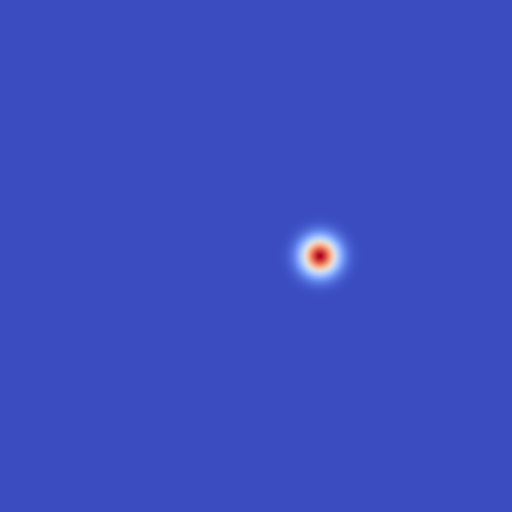} &
    \includegraphics[height=2.5cm,trim={0.4cm 0.4cm 0.4cm 0.4cm},clip]{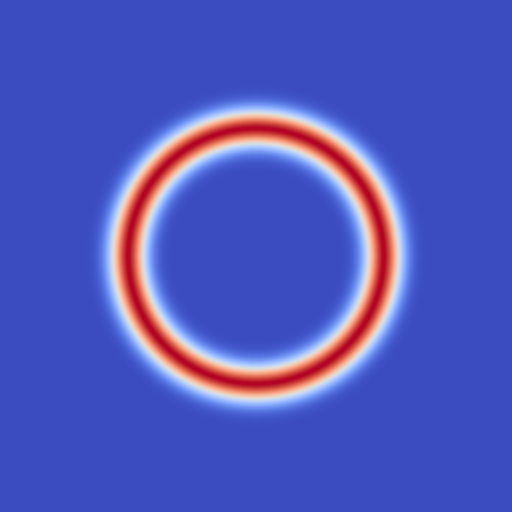} &
    \includegraphics[height=2.5cm,trim={0.4cm 0.4cm 0.4cm 0.4cm},clip]{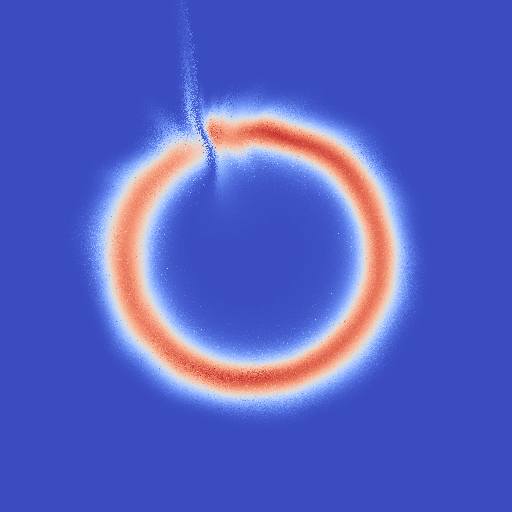} &
    \includegraphics[height=2.5cm,trim={0.4cm 0.4cm 0.4cm 0.4cm},clip]{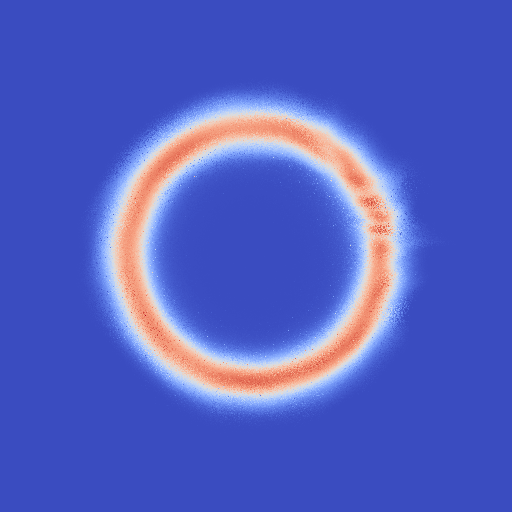} \\
	
	\bottomrule
    
    \end{tabular}}
  \caption{Comparison between conditional affine flows of \citep{atanov2019semi,lu2019structured} and our conditional non-linear (Cond NL) flows. We see that the conditional affine flows cannot fully capture ``ring''-like conditional distributions (note the discontinuity at the top), while our conditional non-linear flows does not have such discontinuities.}
   
  \label{fig:cnlsq_circle}
\end{figure*}

\begin{figure*}[h]
  
  \centering
  \resizebox{\textwidth}{!}{\begin{tabular}{ c@{\hskip 0.5cm}c@{\hskip 0.5cm}c@{\hskip 0.5cm}c@{\hskip 0.5cm}c@{\hskip 0.5cm}c }
  
	\toprule
	 $p(\text{y} | \text{x})$ & MoG, $\text{M}=4$ & MoG, $\text{M}=8$ & MoG, $\text{M}=32$ & MoG, $\text{M}=64$ & \textbf{Our Cond NL Flow} \\
	\midrule
	
	\includegraphics[height=2.5cm,trim={0.4cm 0.4cm 0.4cm 0.4cm},clip]{images/res/cnlsq/gt_circle_cond_nlls_1.png} &
    \includegraphics[height=2.5cm,trim={0.4cm 0.4cm 0.4cm 0.4cm},clip]{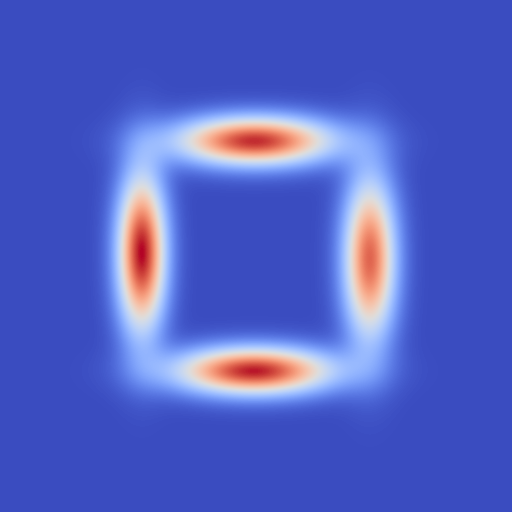} &
    \includegraphics[height=2.5cm,trim={0.4cm 0.4cm 0.4cm 0.4cm},clip]{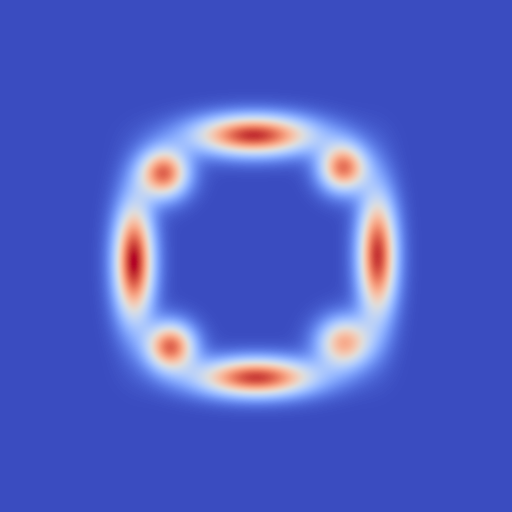} &
    \includegraphics[height=2.5cm,trim={0.4cm 0.4cm 0.4cm 0.4cm},clip]{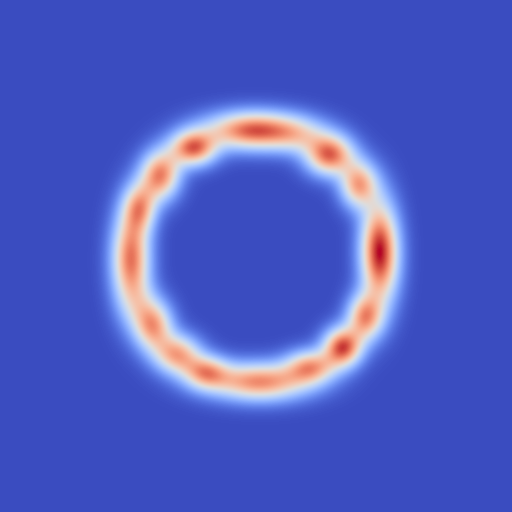} &
    \includegraphics[height=2.5cm,trim={0.4cm 0.4cm 0.4cm 0.4cm},clip]{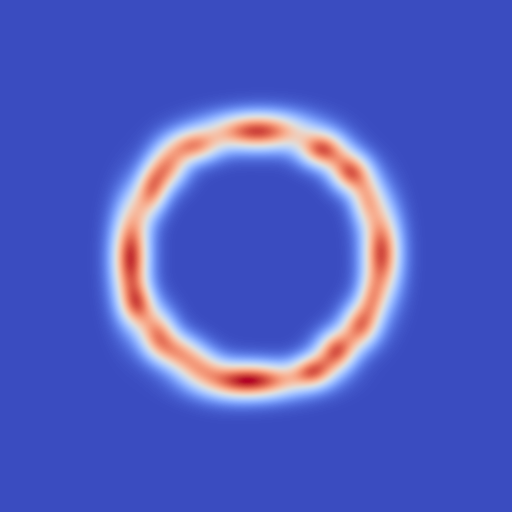} &
	\includegraphics[height=2.5cm,trim={0.4cm 0.4cm 0.4cm 0.4cm},clip]{images/res/cnlsq/circle_cond_nlls_nlsq1.png} \\
	
	\bottomrule
    
    \end{tabular}}
  \caption{Comparison between our conditional non-linear (Cond NL) flows and a Mixture of Gaussians (MoG) model. We see that even with 64 mixture components, the learnt density is not smooth in comparison to our conditional non-linear flows.}
   
  \label{fig:mog_circle}
\end{figure*}

We compare conditional affine flows of \citep{atanov2019semi,lu2019structured} and our conditional non-linear (Cond NL) flows in \autoref{fig:cnlsq_mog} and \autoref{fig:cnlsq_circle}. We plot the conditional distribution $p(\text{y} | \text{x})$ and the corresponding condition $\text{x}$ in the second and first columns. We use 8 and 16 layers of flow in case of the densities in  \autoref{fig:cnlsq_mog} and \autoref{fig:cnlsq_circle} respectively. We see that the estimated density by the conditional affine flows of \citep{atanov2019semi,lu2019structured} contains distinctive ``tails'' in case of \autoref{fig:cnlsq_mog} and discontinuities in case of \autoref{fig:cnlsq_circle}. In comparison our conditional non-linear flows does not have distinctive ``tails'' or discontinuities and is able to complex capture the multi-modal distributions better. Note, the ``ring''-like distributions in \autoref{fig:cnlsq_circle} cannot be well captured by more traditional methods like Mixture of Gaussians. We see in \autoref{fig:mog_circle} that even with 64 mixture components, the learnt density is not smooth in comparison to our conditional non-linear flows. This again demonstrates the advantage of our conditional non-linear flows.

\section*{Appendix C. Additional Details of Our Model Architectures}
Here, we provide details of the model architectures used across the three datasets used in the main paper.

\myparagraph{MNIST Sequences.} We use the same model architecture as in \cite{bhattacharyya2018accurate}. The LSTM condition encoder on the input sequence $\text{x}$, the LSTM recognition network $q_{\theta}$ and the decoder LSTM network has 48 hidden neurons each. Also as in \cite{bhattacharyya2018accurate}, we use  a 64 dimensional latent space.

\myparagraph{Stanford Drone.} Again, we use the same model architecture as in \cite{bhattacharyya2018accurate} except for the CNN encoder. The LSTM condition encoder on the input sequence $\text{x}$ and the decoder LSTM network has 64 hidden neurons each. The LSTM recognition network $q_{\theta}$ has 128 hidden neurons. Also as in \cite{bhattacharyya2018accurate}, we use  a 64 dimensional latent space. Our CNN encoder has 6 convolutional layers of size 32, 64, 128, 256, 512 and 512. We predict the attention weights on the final feature vectors using the encoding of the LSTM condition encoder. The attention weighted feature vectors are passed through a final fully connected layer to obtain the final CNN encoding. Furthermore, we found it helpful to additionally encode the past trajectory as an image (as in \citep{pajouheshgar2018back}) as provide this as an additional channel to the CNN encoder.

\myparagraph{HighD.}  We use the same model architecture with both the CVAE and CF-VAE models. As in the Stanford drone dataset, we use LSTM condition encoder on the input sequence $\text{x}$ and the decoder LSTM network with 64 hidden neurons each and the LSTM recognition network $q_{\theta}$ with 128 hidden neurons. The contextual information of interacting traffic participants are encoded into a spatial grid tensor of size 13$\times$3 (see Section 3.2 of the main paper). We use a CNN with 5 layers of sizes 64, 128, 256, 256 and 256 to extract contextual features.


\section*{Appendix D. Details of the mixture of Gaussians (MoG) baseline}

In the main paper, we include results on the MNIST Sequence and Stanford Drone dataset with a Mixture of Gaussians (MoG) prior. In detail, instead of a normalizing flow, we set the prior to a MoG form,
\begin{align}\label{eq:mog}
p_{\xi}(\text{z}|\text{x}) = \sum\limits_{i=1}^{M} p(\text{c}_{i} | \text{x}) \mathcal{N}(\text{z}; \mu_{i}, \sigma_{i} | \text{x}).
\end{align}

We use a simple feed forward neural network that takes in the condition $\text{x}$ (see Section 3.4 of the main paper) and predicts the parameters of the MoG, $\xi = \left\{ \text{c}_{1}, \mu_{1}, \sigma_{1}, \cdots, \text{c}_{M}, \mu_{M}, \sigma_{M} \right\}$. Note, to ensure a reasonable number of parameters, we consider spherical Gaussians. Similar to (5) in the main paper, the ELBO can be expressed as,

\begin{align}\label{eq:mog_elbo}
\log(p_{\theta}(\text{y}|\text{x})) \geq \mathbb{E}_{q_{\phi}(\text{z}|\text{x},\text{y})} \log(p_{\theta}(\text{y}|\text{z},\text{x})) + \mathcal{H}(q_{\phi}) + \mathbb{E}_{q_{\phi}(\text{z}|\text{x},\text{y})}\log(p_{\xi}(\text{z}|\text{x})).
\end{align}

Note that we fix the entropy of the posterior distribution $q_{\phi}$ for stability

\section*{Appendix E. Additional Evaluation on the MNIST Sequence Dataset}
Here, we perform a comprehensive evaluation using the MoG prior with varying mixture components. Moreover, we experiment with a CVAE with unconditional non-linear flow based prior (NL-CVAE). We report the results in \autoref{tab:mnistseq_mog}.

\begin{table}[h]
\centering
	\begin{tabular}{lc}
	\toprule
	Method & -CLL $\downarrow$ \\
	\midrule
	NL-CVAE & 107.6 \\
	CVAE ($M=1$) \citep{sohn2015learning} & 96.4\\
	\midrule
	MoG-CVAE, $M=2$ & 85.3\\
	MoG-CVAE, $M=3$ & 84.6\\
	MoG-CVAE, $M=4$ & 85.7\\
	MoG-CVAE, $M=5$ & 86.3\\
	
	\midrule
	CF-VAE + pR & \textbf{74.9}\\
	\bottomrule
	\end{tabular}
\caption{Evaluation on MNIST Sequences (CLL: lower is better).}
\label{tab:mnistseq_mog}
\end{table}

 As mentioned in the main paper, we see that the MoG-CVAE outperforms the plain CVAE. This again reinforces our claim that the standard Gaussian prior induces a strong model bias. We see that using $M=3$ components with the variance of the posterior distribution fixed to $\text{C}=0.2$ leads to the best performance. This is expected as 3 is the most frequent number of possible strokes in the MNIST Sequence dataset. Also note that the results with the MoG prior are also relatively robust across $\text{C}=\left[ 0.05, 0.2 \right]$ as we learn the variance of the prior (see the section above). Finally, our CF-VAE + pR still significantly outperforms the MoG-CVAE (74.9 vs 84.6). This is expected as normalizing flows are more powerful compared to MoG at learning complex multi-modal distributions \citep{kingma2018glow} (also see \autoref{fig:mog_circle}).
 
 We also see that using an unconditional non-linear flow based prior actually harms performance (107.6 vs 96.4). This is because the latent distribution is highly dependent upon the condition. Therefore, without conditioning information the non-linear conditional flow learns a global representation of the latent space which leads to out-of-distribution samples at prediction time.


\section*{Appendix F. Evaluation of the Robustness of the Top n\% Metric}
We use two simpler uniform ``Shotgun'' baselines to study the robustness of the Top n\% metric against random guessing. In particular, we consider the ``Shotgun''-u$90^{\circ}$ and ``Shotgun''-u$135^{\circ}$ baselines which: given a budget of N predictions, it uniformly distributes the predictions between $(-90^{\circ},90^{\circ})$ and $(-135^{\circ},135^{\circ})$ respectively of the original orientation and using the velocity of the last time-step. In \autoref{tab:stanford_drone_randguess} we compare the Top 1 (best guess) to Top 10\% metric with N$={50,100,500}$ predictions.

\begin{table*}[h]
\centering
\begin{tabular}{lccccc}
\toprule
Method & K & Error $@$ 1sec & Error $@$ 2sec & Error $@$ 3sec & Error $@$ 4sec \\
\midrule 
&&\multicolumn{4}{c}{Top 1 (Best Guess)} \\
\cline{3-6}
\rule{0pt}{3ex}
``Shotgun''-u$90^{\circ}$ & 50  & 0.9 & 1.9 & 3.1 & 4.4\\
``Shotgun''-u$90^{\circ}$ & 100 & 0.9 & 1.9 & 3.0 & 4.3\\
``Shotgun''-u$90^{\circ}$ & 500 & 0.9 & 1.9 & 3.0 & 4.3\\
\midrule 
&&\multicolumn{4}{c}{Top 10\%} \\
\cline{3-6}
\rule{0pt}{3ex}
``Shotgun''-u$90^{\circ}$ & 50  & 1.2 & 2.5 & 3.9 & 5.4\\
``Shotgun''-u$90^{\circ}$ & 100 & 1.2 & 2.5 & 3.9 & 5.4\\
``Shotgun''-u$90^{\circ}$ & 500 & 1.2 & 2.5 & 3.9 & 5.4\\
\bottomrule
\rule{0pt}{3ex}
&&\multicolumn{4}{c}{Top 1 (Best Guess)} \\
\cline{3-6}
\rule{0pt}{3ex}
``Shotgun''-u$135^{\circ}$ & 50  & 0.9 & 2.0 & 3.1 & 4.5\\
``Shotgun''-u$135^{\circ}$ & 100 & 0.9 & 1.9 & 3.0 & 4.3\\
``Shotgun''-u$135^{\circ}$ & 500 & 0.9 & 1.9 & 3.0 & 4.2\\
\midrule 
&&\multicolumn{4}{c}{Top 10\%} \\
\cline{3-6}
\rule{0pt}{3ex}
``Shotgun''-u$135^{\circ}$ & 50  & 1.4 & 2.9 & 4.5 & 6.2\\
``Shotgun''-u$135^{\circ}$ & 100 & 1.4 & 2.9 & 4.5 & 6.2\\
``Shotgun''-u$135^{\circ}$ & 500 & 1.4 & 2.9 & 4.5 & 6.2\\
\bottomrule
\end{tabular}
\caption{Five fold cross validation on the Stanford Drone dataset. Euclidean error at ($\nicefrac{1}{5}$) resolution.}
\label{tab:stanford_drone_randguess}
\end{table*}

We see that in case of both the ``Shotgun''-u$90^{\circ}$ and ``Shotgun''-u$135^{\circ}$ baselines, the Top 1 (best guess) metric improves with increasing number of guesses. This effect is even more pronounced in case of the ``Shotgun''-u$135^{\circ}$ baseline as the random guesses are distributed over a larger spatial range. In contrast, the Top 10\% metric remains remarkably stable. This is because, in order to improve the Top 10\% metric, random guessing is not enough -- the predictions have to be on the correct modes. In other words, the only way to improve the Top 10\% metric is move random predictions to any of the correct modes.
 
 
\section*{Appendix G. Qualitative Examples on the HighD Dataset}

\begin{figure*}[h]
  
  \centering
  \resizebox{\textwidth}{!}{\begin{tabular}{ c@{\hskip 0.5cm}c@{\hskip 0.5cm}c}
	\toprule
	Groundtruth samples & CVAE samples & Our CF-VAE  + $\left\{\text{pR,cR}\right\}$ samples \\
	\midrule
    \includegraphics[height=5cm,trim={1cm 4cm 4cm 3.5cm},clip]{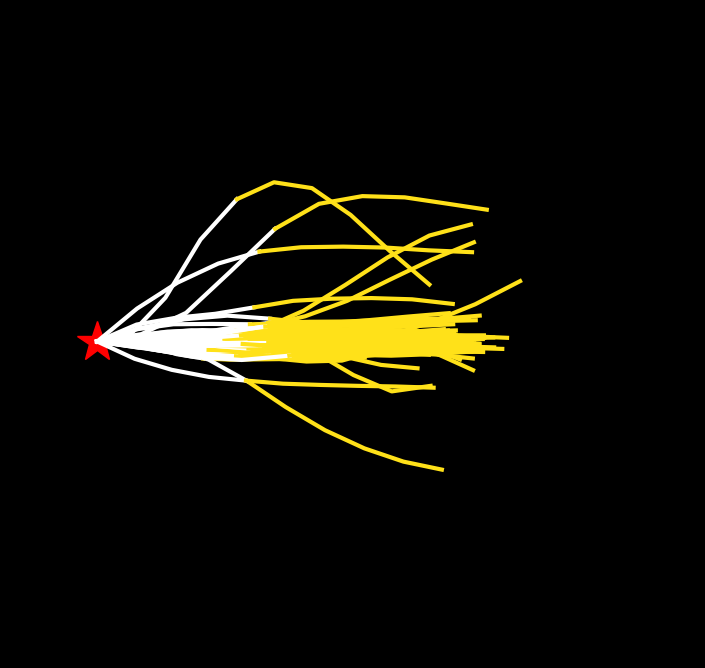} &
    \includegraphics[height=5cm,trim={1cm 4cm 4cm 3.5cm},clip]{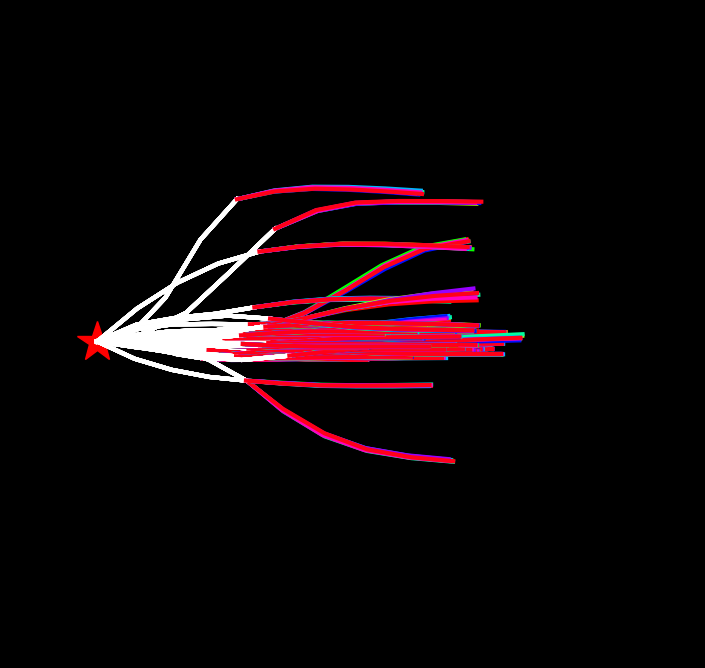} &
	\includegraphics[height=5cm,trim={1cm 4cm 4cm 3.5cm},clip]{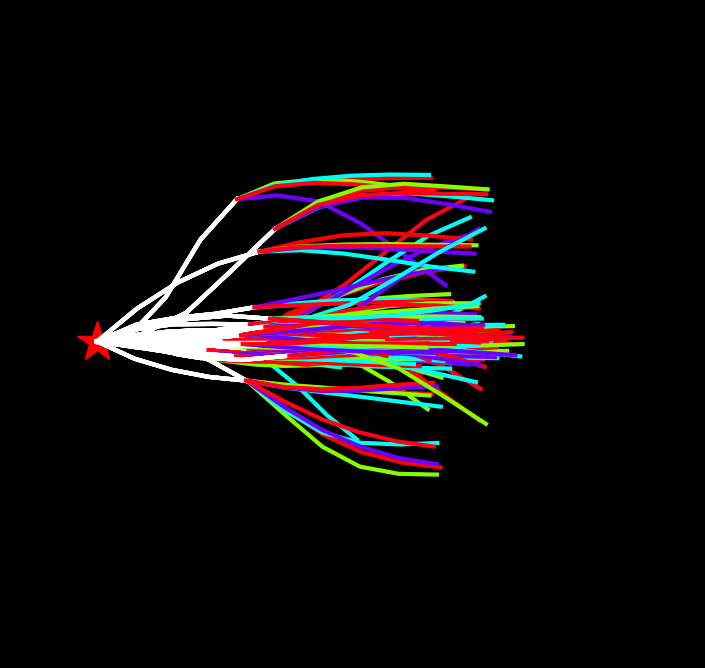} \\
    \bottomrule
    \end{tabular}}
  \caption{Predictions on the HighD dataset. Left: 128 random samples from the HighD test set (in yellow). Middle: CVAE predictions (5 samples per test set example). Right: Our CV-VAE  + $\left\{\text{pR,cR}\right\}$ predictions (5 samples per test set example). While the predictions by the CVAE are linear continuations, our CF-VAE sample predictions are much more diverse and cover events like lane changes e.g.\ top most sample track from the test set. }
   
  \label{fig:highD}
\end{figure*}

We show qualitative examples on the HighD dataset in \autoref{fig:highD}. In the left of \autoref{fig:highD} we show 128 random samples from the HighD test set. In the middle we show predictions on these samples by the CVAE (with cyclic Kl annealing \citep{liu2019cyclical}). We see that even with cyclic KL annealing, we observe posterior collapse. All samples have been pushed towards the mean and the variance in the 5 samples per test set example is minimal. E.g. note the top most sample track from the test set in \autoref{fig:highD} (left). All CVAE sample predictions are a linear continuation of the trajectory (continuing on the same lane), while there is in fact a turn (change of lanes). In contrast, our CF-VAE + $\left\{\text{pR,cR}\right\}$ sample predictions are much more diverse and cover such eventualities. This also shows that our CF-VAE + $\left\{\text{pR,cR}\right\}$ does not suffer from such posterior variable collapse.

\end{document}